\documentclass[sigconf]{acmart}

\usepackage{url}            %
\usepackage{booktabs}       %
\usepackage{amsfonts}       %
\usepackage{nicefrac}       %
\usepackage{microtype}      %
\usepackage{xcolor}         %

\usepackage{wrapfig,lipsum,booktabs}
\usepackage{microtype}
\usepackage{graphicx}
\usepackage{booktabs} %
\usepackage{bm}
\usepackage{subfig}
\usepackage[inline, shortlabels]{enumitem}
\usepackage{pspicture}
\usepackage{textcomp}
\usepackage{booktabs}
\usepackage{caption}
\usepackage{algorithm,algorithmicx,algpseudocode}
\usepackage{gensymb}
\usepackage{verbatim}
\usepackage{euscript}
\usepackage{float}
\usepackage{upgreek}
\usepackage{amsmath,amsthm,amsfonts,mathtools}

\usepackage{threeparttable}
\usepackage{graphicx}
\usepackage{bbm}
\usepackage{multirow}

\newcommand{\norm}[1]{\left\lVert#1\right\rVert}

\DeclarePairedDelimiter\set\{\}

\DeclareMathOperator{\E}{\mathbb{E}} %
\DeclareMathAlphabet{\pazocal}{OMS}{zplm}{m}{n} %

\newcommand{\Tau}{\mathcal{T}}
\newcommand{\half}{\textstyle\frac{1}{2}}

\usepackage[utf8]{inputenc} %
\usepackage[T1]{fontenc}    %

\begin{document}
\fancyhead{}
\title{Representation Learning via Variational Bayesian Networks}
\author{Oren Barkan}
\authornote{Authors contributed equally to this research.}
\affiliation{%
  \institution{The Open University}
  \country{Israel}
}
\author{Avi Caciularu}
\authornotemark[1]
\affiliation{%
  \institution{Bar-Ilan University}
  \country{Israel}
}
\author{Idan Rejwan}
\authornotemark[1]
\affiliation{%
  \institution{Tel Aviv Univeristy}
  \country{Israel}
}

\author{Ori Katz}
\affiliation{%
  \institution{Technion}
  \country{Israel}
}

\author{Jonathan Weill}
\affiliation{%
  \institution{Tel Aviv Univeristy}
  \country{Israel}
}

\author{Itzik Malkiel}
\affiliation{%
  \institution{Tel Aviv University}
  \country{Israel}
}

\author{Noam Koenigstein}
\affiliation{%
  \institution{Tel Aviv University}
  \country{Israel}
}

\begin{abstract}

We present Variational Bayesian Network (VBN) - a novel Bayesian entity representation learning model that utilizes hierarchical and relational side information and is particularly useful for modeling entities in the ``long-tail'', where the data is scarce. VBN provides better modeling for long-tail entities via two complementary mechanisms: First, VBN employs informative hierarchical priors that enable information propagation between entities sharing common ancestors. Additionally, VBN models explicit relations between entities that enforce complementary structure and consistency, guiding the learned representations towards a more meaningful arrangement in space. Second, VBN represents entities by densities (rather than vectors), hence modeling uncertainty that plays a complementary role in coping with data scarcity. Finally, we propose a scalable Variational Bayes optimization algorithm that enables fast approximate Bayesian inference. We evaluate the effectiveness of VBN on linguistic, recommendations, and medical inference tasks. Our findings show that VBN outperforms other existing methods across multiple datasets, and especially in the long-tail.

\end{abstract}




\maketitle

\section{Introduction}
\label{sec:intro}
Entity representation learning is an active research field with applications in 
natural language understanding (NLU) \cite{mikolov2013distributed,pennington2014glove,bojanowski2017enriching, barkan2020scalable}, recommender systems \cite{Koren_MF, salakhutdinov2008bayesian,barkan2020attentive,i2v,barkan2020explainable,anchors,nam}, medical informatics \cite{lin2019projection}, and more.  %
In the last decade, a variety of non-contextualized representation learning models were developed \cite{mikolov2013distributed,salakhutdinov2008bayesian,zhang-etal-2014-word,3327687,vilnis2014word}. 
These models learn representations using large datasets of co-occurrences in a self-supervised fashion.
However, these datasets often incorporate a long-tail of rare entities with very little co-occurrence data. 
In the recommender system community, this situation is known as the "cold-start" problem~\cite{nam,barkan2019cb2cf}, where rare ('cold') entities (e.g., unpopular items or new items that are introduced to the catalog) are often poorly represented due to insufficient statistics. 
In the natural language processing community, where the focus is on learning representations for words and phrases, a common mitigation is to increase the training set size by utilizing increasingly larger corpus e.g., BERT \cite{devlin-etal-2019-bert,liu2019roberta}. 
However, it was shown that even when increasing the amount of co-occurrence data, the existence of rare, out-of-vocabulary entities persists \cite{pinter-etal-2017-mimicking,schick-schutze-2019-attentive,hice2019,schick2020rare}.

A recent attempt proposed a denoising method via fusing several embedding sets for improving the quality of long-tail words \cite{caciularu-etal-2021-denoising}, but it still relies on the basic ability to learn from co-occurrence statistics. Moreover, in other applications such as recommender systems, medical informatics, etc., co-occurrence data is limited and pre-trained models are generally not in existence. Therefore, in these domains, finetuning pre-trained representations is usually impossible, or extremely limited at best.

While increasing the dataset size is often impossible, other side information on entities might be available and leveraged for mitigating the cold-start problem. %
For example, in many domains, entities follow a well-defined taxonomy that ties related entities to each other (e.g., \emph{genre}$\rightarrow$\emph{artist}$\rightarrow$\emph{song}).
Additionally, relational information that defines a relation or a particular type of ``connection'' between two entities (e.g., \emph{warm} and \emph{cold} are antonyms, \emph{page} and \emph{cover} are meronyms of \emph{book}), can be utilized as well. 
Therefore, in this work, we introduce the Variational Bayesian Network (VBN) - a novel Bayesian representation learning model that incorporates external hierarchical and relational side information, in addition to co-occurrence relations. VBN is particularly useful in small data scenarios and for modeling long-tail entities. Our contributions are as follows: 
\begin{itemize}
    \item We introduce the novel VBN objective that facilitates joint modeling of three types of complementary relations:
(a) Explicit hierarchical relations via a network of informative priors that enables information propagation between entities to improve representation of entities in the long-tail. 
(b) Explicit relational information (e.g. antonyms, meronyms, etc.) that enforces structure and consistency between related entities. (c) Implicit relations (co-occurrences) that capture semantic and syntactic information between entities.
\item We present a tractable yet scalable Variational Bayes (VB) optimization algorithm that maps the entities into \emph{probability densities} (approximate posteriors).
The Bayesian approach is complementary to the aforementioned hierarchical priors and enables better treatment of uncertainty in long-tail entities. This is in contrast to point estimate solutions that treat the latent variables as \emph{parameters}.
\item While our proposed VB algorithm produces a fully factorized posterior approximation, the inference phase still involves intractable integration. Our third contribution is an analytical approximation of the posterior predictive integral that enables fast Bayesian inference.
\end{itemize}

 We demonstrate the effectiveness of VBN on NLU, recommendations, and medical inference tasks. VBN is shown to significantly outperform a variety of state-of-the-art methods across all datasets, and especially in the long-tail.

\section{Related Work}
\label{sec:related}
Incorporating external side information in representation learning has been studied extensively \cite{yu-dredze-2014-improving,wu2016collaborative, li-etal-2016-joint,cvae,wang2019neural,10.1145/3383313.3412239,barkan2020cold,barkan2019cb2cf,nam,anchors}.
Retrofitting \cite{faruqui-etal-2015-retrofitting} is a post-processing technique that was introduced in order to refine pretrained word representations using relational information from semantic lexicons. %
In \cite{bollegala2016joint}, the authors proposed methods to learn word representations subject to relational constraints but without the utilization of their hierarchical structure. 
Recently, contextualized word embedding models were expanded to include external linguistic information during pretraining \cite{huang-etal-2019-glossbert,levine2019sensebert,Peters2019KnowledgeEC,arora-etal-2020-learning,caciularu2021cross}. Yet, these models require massive amounts of data to be effective. 
In contrast, VBN models both hierarchical and relational information, during learning, and performs well in small data scenarios.

Bayesian representation learning models were previously proposed in \cite{salakhutdinov2008bayesian,zhang-etal-2014-word,bravzinskas2017embedding, barkan2017bayesian, vilnis2014word,paquet2013,koenigstein13,barkan-etal-2020-bayesian}, and representation learning using graphical models was presented in \cite{3327687,tifrea2018poincare,vilnis2014word,3186150,3061142}. Much of these works revolve around NLU and recommender systems.
While the aforementioned works do not make explicit use of external side information, one may propose to apply the method from \cite{faruqui-etal-2015-retrofitting} to enhance them with word taxonomy. Our evaluations show that VBN outperforms this alternative.

\section{Variational Bayesian Networks}
\label{sec:vbnets}
VBN is a probabilistic graphical model in which an entity $i$ can appear either as a leaf node, a parent (internal) node, or both.  
Let $\mathcal{I}=\{i\}_{i=1}^{N_W}$ be a set of $N_w$ entities (for simplicity, entities are indexed by numbers). 
Entity nodes are unobserved variables (representations) that are being learned. Specifically, each entity $i$ is associated with two \emph{leaf} representations
$\mathbf{u}_{i},\mathbf{v}_{i}\in\mathbb{R}^t$ (similar to the \textit{context} and \textit{target} representations in log bi-linear modeling \cite{mikolov2013distributed}). Similarly, $\mathbf{h}_i^u,\mathbf{h}_i^v\in\mathbb{R}^t$ are the \emph{parent} representations of entity $i$. Hence, if entity $i$ is a parent of entity $j$, then the nodes $\mathbf{h}_i^u$ and $\mathbf{h}_i^v$ serve as parents to the nodes $\mathbf{u}_{j}$ and $\mathbf{v}_{j}$, respectively. In addition, every node can have multiple parents and children. Thus, we further define $\pi_i,\omega_i\subset\mathcal{I}$ as the sets of parents and children entities of entity $i$, respectively.

\begin{figure*}[t!]
      \centering
    \includegraphics[scale=0.52]{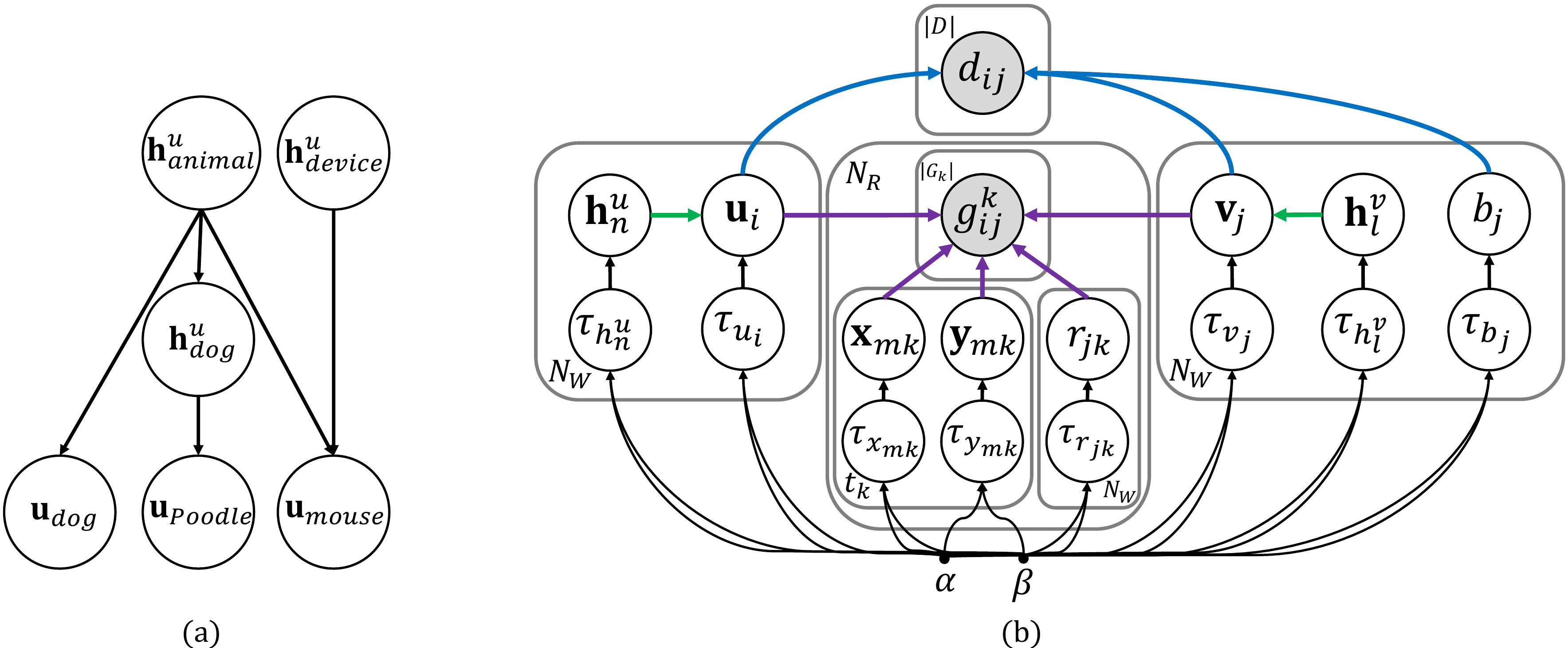}
\caption{\textbf{(a)} A toy VBN example for the entities \emph{dog}, \emph{Poodle}, \emph{mouse} and their parents. Only the $u$ part is presented (the $v$ part is symmetric. See Section~\ref{sec:vbnets} for details). \textbf{(b)} VBN's graphical model. Gray variables are observed (data), whereas white variables are unobserved (learned). $d_{ij}$ are the co-occurrence data, explained (blue arrows) by the leaf entity representations $\mathbf{u}_i,\mathbf{v}_j$ and bias $b_j$ (Section \ref{subsec:cooccurrence}). $g_{ij}^k$ are the explicit relations data, explained (purple arrows) by $\mathbf{u}_i$ and $\mathbf{v}_j$, the linear mappings induced by $\mathbf{x}_{mk}$ and $\mathbf{y}_{mk}$, and the bias term $r_{jk}$ (Section \ref{subsec:explicit}). $\mathbf{h}_{n}^u$ and $\mathbf{h}_{l}^v$ are the parent entity representations that form hierarchical priors (green arrows) over $\mathbf{u}_i$ and $\mathbf{v}_j$, respectively (Section \ref{subsec:Hier}). $\tau_{u_i},\tau_{v_j},\tau_{x_{mk}},\tau_{y_{mk}},\tau_{h^u_n},\tau_{h^v_l},\tau_{b_j},\tau_{r_{jk}}$ are the learned precision variables with Gamma hyperprior, where $\alpha$ and $\beta$ are the shape and rate hyperparameters, respectively.}
\label{fig:fig1}\vspace{4mm}
\end{figure*}

Figure \ref{fig:fig1}(a) presents a toy example of VBN for the entities \emph{dog}, \emph{Poodle} and \emph{mouse}, and their parents (note that only the `$u$' part of the graph is shown. The `$v$' part is symmetric). In this example, $\mathbf{h}_{animal}^u$ is the parent of $\mathbf{h}_{dog}^u$, $\mathbf{u}_{dog}$ and $\mathbf{u}_{mouse}$. It is important to distinguish between $\mathbf{u}_{dog}$ which represents the actual \textbf{word} \emph{dog}, and  $\mathbf{h}_{dog}^u$ that represents the \textbf{category} \emph{dog}, which is in turn a parent of $\mathbf{u}_{Poodle}$ that represents the word \emph{Poodle} (dog breed). Note that $\mathbf{u}_{mouse}$ has two parents: $\mathbf{h}_{animal}^u$ and $\mathbf{h}_{device}^u$, as the word \emph{mouse} is ambiguous. Further note that the word representations of the entities \emph{animal} and \emph{device} are given by the leaf nodes $\mathbf{u}_{animal}$ and $\mathbf{u}_{device}$, respectively (not shown).

\subsection{Hierarchical Relations}
\label{subsec:Hier}
VBN models three types of relations between entities: hierarchical, explicit, and co-occurrence. 
The model utilizes hierarchical information via informative priors. For example, Fig.~\ref{fig:fig1}(a) presents the hierarchy \textit{animal $\rightarrow$ dog $\rightarrow$ Poodle}. In music recommendations, taxonomy exhibits the following hierarchy: \textit{genre $\rightarrow$ subgenre $\rightarrow$ artist $\rightarrow$ song}, where each parent entity is used as a prior over its child entity. Let $\mathbf{U}=\{\mathbf{u}_i\}_{i\in\mathcal{I}}$, $\mathbf{H}^u=\{\mathbf{h}_i^u\}_{i\in\mathcal{I}}$, and denote $\mathbf{s}_i^u \triangleq |\pi_i|^{-1}\sum_{n\in\pi_i}\mathbf{h}_n^u$, 
if ${\pi_i \neq \emptyset}$, otherwise ${\mathbf{s}_i^u}$ is set to ${\mathbf{0}}$. We assume Normal-Gamma hierarchical priors:
\begin{equation}
\label{eq:h_u}
\begin{split}
   p(\mathbf{U},\mathbf{H}^u,\bm{\Tau}^{u},\bm{\Tau}^{h^u}|\mathcal{H})=\textstyle\prod_{i\in\mathcal{I}}& \mathcal{N}(\mathbf{u}_i;\mathbf{s}_i^u,\tau_{u_i}^{-1}\mathbf{I})\mathcal{G}(\tau_{u_i};\alpha,\beta)\\&
   \mathcal{N}(\mathbf{h}_i^u;\mathbf{s}_i^u,\tau_{h_i^u}^{-1}\mathbf{I})\mathcal{G}(\tau_{h_i^u};\alpha,\beta),
\end{split}
\end{equation}
where $\bm{\Tau}^{u}=\{\tau_{u_i}\}_{i\in\mathcal{I}}$ and $\bm{\Tau}^{h^u}=\{\tau_{h_i^u}\}_{i\in\mathcal{I}}$ ($\tau_{u_i},\tau_{h_i^u}\in\mathbb{R}$) are the precision variables that follows the Gamma hyperpriors
with shape and rate hyperparameters  $\mathcal{H}=\{\alpha,\beta\}$.   
The hierarchical priors in Eq.~\ref{eq:h_u} enforce entity representations to be closer to their parents, in terms of $L^2$ distance, where the precision variables control the strength of this constraint. This enables child nodes to naturally ``fallback'' on to their parents in case of insufficient statistics. As we will see, this unique feature improves the representation of long-tail entities. Further note that the formulation in Eq.~\ref{eq:h_u} supports hierarchical relations of arbitrary depths. Finally, $\mathbf{V}, \mathbf{H}^v, \bm{\Tau}^{h^v}, \bm{\Tau}^{v}$ and $\mathbf{s}_i^v$ together with their priors and hyperpriors are defined in a symmetric manner. The hierarchical relations appear in Fig. \ref{fig:fig1}(b) (green).

We note that in the general case, the hierarchical priors can be modeled via multimodal distributions (e.g., Gaussian Mixture Models). Moreover, various types of content (e.g., image, audio, free text) can be processed by a neural network that predicts the parameters of the distribution. However, we leave the investigation of these extensions  %
for future work.

\subsection{Co-occurrence Relations}
\label{subsec:cooccurrence}
VBN learning is based on modeling co-occurrence relations. These co-occurrences can be
words that appear next to each other, co-consumed items, co-morbid diseases, etc.
Let $I_P=\{(i,j)|$ entity $j$ occurs in the context of entity $i\}$ be the co-occurrence dataset (positive relations). Note that $I_P$ is a multiset, i.e. $(i,j)$ can appear multiple times in $I_P$. %
During training, we subsample \cite{mikolov2013distributed} a new ${I_{P^\text{'}}\subset I_P}$ every epoch. Then, for each positive pair $(i,j)\in I_{P^\text{'}}$, we sample $n$ negative pairs ${\set{(i,z_l)|(i,z_l)\notin I_{P^{\text{'}}}}}_{l=1}^n$ to form a negative multiset $I_N$. Hence, the training set ${I_D={I_{P^\text{'}}}\cup I_N}$ is stochastic.

VBN models co-occurrence relations as a classification problem via a two-point random variable ${d:\mathcal{I}\times\mathcal{I}\rightarrow\{1,-1\}}$ with ${d_{ij}=1}$ if ${(i,j)}\in I_P'$, otherwise ${d_{ij}=-1}$. Let ${\mathbf{D}=\{d_{ij}|(i,j) \in I_D\}}$ be the training data ($\mathbf{D}$ are observed). The likelihood is given by:
\begin{equation}
\label{eq:cooccur-likelihood}
\begin{split}
    p(\mathbf{D}|\mathbf{U},\mathbf{V},\mathbf{B}) &=\textstyle\prod_{(i,j)\in I_D} \sigma(d_{ij}(\mathbf{u}_i^T\mathbf{v}_j+b_j)),
\end{split}
\end{equation}
where ${\sigma(x)=1/(1+e^{-x})}$.  ${\mathbf{B}=\{b_i\}_{i\in\mathcal{I}}}$  ${(b_i\in\mathbb{R})}$ are bias variables with Normal-Gamma priors:
\begin{equation}
\label{eq:bias}
\begin{split}
  p(\mathbf{B},\bm{\Tau}^b|\mathcal{H})&=\textstyle\prod_{i\in\mathcal{I}}\mathcal{N}(b_i;0,\tau_{b_i}^{-1})\mathcal{G}(\tau_{b_i};\alpha,\beta).
\end{split}
\end{equation}
Equation \ref{eq:cooccur-likelihood} explains the observed variables ${d_{ij}\in \mathbf{D}}$ (the data) by the unobserved variables ${\mathbf{u}_i, \mathbf{v}_j}$ and ${b_j}$, as depicted in Fig. \ref{fig:fig1}(b) (in blue).

\subsection{Explicit Relations}
\label{subsec:explicit}
VBN further utilizes explicit semantic relations (e.g., \emph{antonyms}, \emph{meronyms}, etc.) to enforce additional structure in the latent space and yield more meaningful representations.
To this end, VBN learns representations for any explicit relation and compels entities that share the same relation to ``adhere'' as we explain next.  

We denote by ${i\xrightarrow{k}j}$ the fact that entities $i$ and $j$ share a directed relation $k$. For undirected relations, ${i\xrightarrow{k}}j \land j{\xrightarrow{k}i}$ holds.
A dataset of `type $k$ relations' is constructed in a stochastic manner similarly to the procedure described in Section \ref{subsec:cooccurrence}: Let ${I_P^k=\set{(i,j)|i\xrightarrow{k}j}}$ be the set of positive pairs. For each ${(i,j)\in I_P^k}$, we uniformly sample $n$ negative pairs ${\set{(i,z_l)|(i,z_l)\notin I_P^k}}_{l=1}^n$ to form the negative multi-set ${I_N^k}$. Finally, ${I_G^k=I_P^k \cup I_N^k}$ is the sampled dataset. This procedure is repeated every epoch by sampling a new ${I_N^k}$, while keeping ${I_P^k}$ fixed throughout the training process.

Similarly to co-occurrences, we model ${i\xrightarrow{k}j}$ via a two-point random variable
${g^k:\mathcal{I}\times\mathcal{I}\rightarrow\set{1,-1}}$ with ${g^k_{ij}=1}$ iff ${(i,j)\in I_P^k}$, and ${\mathbf{G}^k=\set{g^k_{ij}|(i,j) \in I_P^k}}$ ($\mathbf{G}^k$ are observed). For $N_R$ different types of explicit relations, we define ${\mathbf{G}=\{\mathbf{G}^k\}_{k=1}^{N_R}}$, and the likelihood is given by:
\begin{equation}
\label{eq:relation-likelihood}
\begin{split}
    p(\mathbf{G}|\mathbf{U},\mathbf{V},\mathbf{W},\mathbf{R}) =\textstyle\prod_{k=1}^{N_R}\textstyle\prod_{(i,j)\in I_G^k} \sigma(g^k_{ij}(\mathbf{u}_i^T\mathbf{W}_k\mathbf{v}_j+r_{jk})),
\end{split}
\end{equation}
with $\mathbf{W}=\{\mathbf{W}_k\}_{k=1}^{N_R}$ and $\mathbf{W}_k=\mathbf{X}_k\mathbf{Y}_k^T$, where $\mathbf{X}_k$ and $\mathbf{Y}_k$ are low-rank matrices in $\mathbb{R}^{t \times t_k}$ whose columns are unobserved random vectors $\mathbf{x}_{mk}\in \mathbb{R}^t$ and $\mathbf{y}_{mk}\in \mathbb{R}^t$ ($1\leq m\leq t_k$), respectively, and $\mathbf{R}=\{r_{ik}\}_{1\leq k \leq N_R,i\in\mathcal{I}}$ are biases, with Normal-Gamma priors as follows:
\begin{equation}
\label{eq:r_prior}
\begin{split}
  p(\mathbf{W},\mathbf{R},\bm{\Tau}^x,\bm{\Tau}^y,\bm{\Tau}^r|\mathcal{H})=\textstyle\prod_{k=1}^{N_R}&\textstyle\prod_{i\in \mathcal{I}}\mathcal{N}(r_{ik};0,\tau_{r_{ik}}^{-1})\mathcal{G}(\tau_{r_{ik}};\alpha,\beta)\\&\textstyle\prod_{m=1}^{t_k}\mathcal{N}(\mathbf{x}_{mk};0,\tau_{x_{mk}}^{-1}\mathbf{I})\mathcal{G}(\tau_{x_{mk}};\alpha,\beta)\\&\quad\quad\quad\mathcal{N}(\mathbf{y}_{mk};0,\tau_{y_{mk}}^{-1}\mathbf{I})\mathcal{G}(\tau_{y_{mk}};\alpha,\beta).
\end{split}
\end{equation}

Equation \ref{eq:relation-likelihood} explains the data ${g_{ij}^k\in \mathbf{G}^k}$ by the unobserved variables ${\mathbf{u}_i}$, $\mathbf{v}_j$, ${\mathbf{X}_k}$, $\mathbf{Y}_k$ and $r_{jk}$, as depicted in Fig. \ref{fig:fig1}(b) (in purple). ${\mathbf{X}_k}$ and ${\mathbf{Y}_k}$ map ${\mathbf{u}_i}$ and ${\mathbf{v}_j}$ to a subspace s.t. if ${i\xrightarrow{k}j}$, then $\mathbf{u}_i^T\mathbf{W}_k\mathbf{v}_j+r_{jk}$ is maximized. Specifically, in the case of undirected relations (e.g., antonyms), $\mathbf{u}_j^T\mathbf{W}_k\mathbf{v}_i+r_{ik}$ is maximized as well. 

\subsection{The Joint Distribution}
\label{subsec:joint distribution}
We denote the data by ${\bm{\mathcal{D}}=\set{\mathbf{D},\mathbf{G}}}$, the unobserved (learned) variables by $\bm{\theta}=\set{\mathbf{U},\mathbf{V},\mathbf{W},\mathbf{B},\mathbf{R},\mathbf{H}^u,\mathbf{H}^v,\bm{\Tau}^u,\bm{\Tau}^v,\bm{\Tau}^x,\bm{\Tau}^y,\bm{\Tau}^b,\bm{\Tau}^r,\bm{\Tau}^{h^u},\bm{\Tau}^{h^v}}$, and the hyperparameters by ${\mathcal{H}=\set{\alpha,\beta}}$. Then, the joint log distribution is $\log p(\bm{\mathcal{D},\bm{\theta}}|\mathcal{H})=\log p(\bm{\mathcal{D}}|\bm{\theta}) + \log p(\bm{\theta}|\mathcal{H})$,
where the joint log likelihood (Eqs.~\ref{eq:cooccur-likelihood},~\ref{eq:relation-likelihood}) is given by:
\begin{equation}
\label{eq:joint likelihood}
\begin{split}
    \log p(\bm{\mathcal{D}}|\bm{\theta})&=\sum_{(i,j)\in I_D} \log \sigma (d_{ij}  (\mathbf{u}_i^T \mathbf{v}_j+b_j)) \\&\quad+ \sum_{k=1}^{N_R} \sum_{(i,j)\in I_G^k} \log \sigma (g^k_{ij}(\mathbf{u}_i^T\mathbf{W}_k\mathbf{v}_j+r_{jk})),
\end{split}
\end{equation}
and the log prior (Eqs.~\ref{eq:h_u},\ref{eq:bias},\ref{eq:r_prior}) is given by:
\begin{equation}
\label{eq:prior}
\begin{split}
\log p(\bm{\theta}|\mathcal{H})=&
-\frac{1}{2} \sum_{i \in \mathcal{I}} \tau_{u_i}\norm{\mathbf{u}_i-\mathbf{s}_i^u}_2^2 
+ \tau_{v_i}\norm{\mathbf{v}_i-\mathbf{s}_i^v}_2^2 
\\&+ \tau_{b_i}b_i^2  +\tau_{h^u_i}\norm{\mathbf{h}_i^u-\mathbf{s}_i^u}_2^2
 +\tau_{h^v_i}\norm{\mathbf{h}_i^v-\mathbf{s}_i^v}_2^2 \\&
+ \sum_{k=1}^{N_R} \tau_{r_{ik}}r^2_{ik} -\frac{1}{2} \sum_{k=1}^{N_R}\sum_{m=1}^{t_k} \tau_{x_{mk}}\norm{\mathbf{x}_{mk}}_2^2 + \tau_{y_{mk}}\norm{\mathbf{y}_{mk}}_2^2 \\&
+ (\alpha + \frac{t}{2} - 1) \left[ \sum_{i \in \mathcal{I}} \log \tau_{u_i}+ \log \tau_{v_i} +\log \tau_{{h^u_i}} +\log \tau_{{h^v_i}}   \right. \\ 
&\left. +\sum_{k=1}^{N_R}\sum_{m=1}^{t_k}\log \tau_{x_{mk}}+\log \tau_{y_{mk}}\right] + \sum_{i \in \mathcal{I}} (\alpha + \frac{1}{2} - 1)[\log \tau_{b_i} \\&+\sum_{k=1}^{N_R}\log \tau_{r_{ik}}] - \beta(\tau_{u_i} + \tau_{v_i} + \tau_{h^u_i} +  \tau_{h^v_i} + \tau_{b_i} +\sum_{k=1}^{N_R}\tau_{r_{ik}}) \\& -\beta \sum_{k=1}^{N_R}\sum_{m=1}^{t_k} \tau_{x_{mk}} + \tau_{y_{mk}} + \text{const}.
\end{split}
\end{equation}

\subsection{VBN Optimization and Inference}
\label{subsec:approximate bayesian inference}
We aim at computing the posterior predictive distribution of ${d_{ij}}$ and ${g^k_{ij}}$.
For brevity, we focus on ${d_{ij}}$, since ${g^k_{ij}}$ is computed in the same manner. The posterior predictive probability of entity $j$ to occur in the vicinity of entity $i$ is written as a marginalization over $\bm{\theta}$ as follows:
\begin{equation}
\label{eq:posterior-predicitive}
    p(d_{ij}=1|\bm{\mathcal{D}}, \mathcal{H})=\int{ \sigma(\mathbf{u}_i^T\mathbf{v}_j+b_j)p(\bm{\theta}|\bm{\mathcal{D}},\mathcal{H})d\bm{\theta}}.
\end{equation}

The posterior ${p(\bm{\theta}|\bm{\mathcal{D}},\mathcal{H})}$ in Eq. \ref{eq:posterior-predicitive} is intractable. Hence, we turn to a VB approximation \cite{bishop2006pattern} of ${p(\bm{\theta}|\bm{\mathcal{D}},\mathcal{H})}$ with a fully factorized distribution ${q(\bm{\theta})=\prod_{z\in\bm{\theta}}q(z)}$. ${q(\bm{\theta})}$ is obtained by minimizing the Kullback-Lieber divergence $D_{KL}(q(\bm{\theta})||p(\bm{\theta}|\bm{\mathcal{D}},\mathcal{H}))$, which is equivalent to the maximization of the variational free energy \cite{bishop2006pattern}: 
$\mathcal{L}(q) =-D_{KL}(q(\bm{\theta})||p(\bm{\theta}|\bm{\mathcal{D}},\mathcal{H}))+\log p(\bm{\mathcal{D}}|\mathcal{H}).$
The maximization of $\mathcal{L}(q)$ is done by following an iterative procedure (which is guaranteed to converge \cite{bishop2006pattern}). At each iteration, for each ${z\in\bm{\theta}}$, update ${q(z)}$ according to the general VB update rule:
\begin{equation}
\label{eq:vb-update-rule}
    q^*(z)=\text{exp}(\mathbb{E}_{q(\bm{\theta}\setminus z)}[\text{log}p(\bm{\theta},\bm{\mathcal{D}}|\mathcal{H})]+const).
\end{equation}

Unfortunately, a straightforward application of the VB update rule from Eq.~\ref{eq:vb-update-rule} is impractical. This happens since the normal priors in Eq.~\ref{eq:prior} are not conjugate to the Bernoulli likelihoods in Eq.~\ref{eq:joint likelihood}, hence the joint distribution $p(\bm{\theta},\bm{\mathcal{D}}|\mathcal{H})$ does not belong to the exponential family. Therefore, to enable conjugacy, we propose to lower-bound the logistic terms in Eq.~\ref{eq:joint likelihood} with Gaussian functions by using the following bound from \cite{jaakkola1997variational}:
\begin{equation}\label{eq:the-bound}
    \log \sigma(x)\geq \frac{x-\delta}{2}-\lambda(\delta)(x^2-\delta^2 )+\log\sigma(\delta),
\end{equation}
where $\delta$ is the variational parameter and $\lambda(\delta)\triangleq\frac{1}{2\delta}(\sigma(\delta)-\frac{1}{2})$. Note that the bound becomes tight for $\delta=x$. We lower-bound each term in the joint log likelihood (Eq.~\ref{eq:joint likelihood}) according to Eq.~\ref{eq:the-bound}, thus introducing a variational parameter $\xi_{ij}$ for any co-occurrence $(i,j)\in I_D$ and $\zeta^k_{ij}$ for any explicit relation $(i,j)\in I_G^k$ for $1 \leq k \leq N_R$ as follows:
\begin{equation}
\label{eq:bounded-likelihood}
    \begin{split}
    \log p(\bm{\mathcal{D}|\theta})&\geq\log\Tilde{p}(\bm{\mathcal{D}|\theta})\\&= \sum_{(i,j)\in I_D}\frac{d_{ij} (\mathbf{u}_i^T\mathbf{v}_j+b_j)-\xi_{ij}}{2}
    \\&
    -\lambda(\xi_{ij})(\mathbf{u}_i^T \mathbf{v}_j \mathbf{v}_j^T \mathbf{u}_i+2\mathbf{u}_i^T \mathbf{v}_j b_j+b_j^2-\xi_{ij}^2 )
    \\&
    +\log\sigma(\xi_{ij}) +\sum_{k=1}^{N_R}\sum_{(i,j)\in I_G^k}\frac{g_{ij}^k(\mathbf{u}_i^T\mathbf{W}_k\mathbf{v}_j+r_{jk})-\zeta^k_{ij}}{2}
    \\&
    -\lambda(\zeta^k_{ij})(\mathbf{u}_i^T\mathbf{W}_k\mathbf{v}_j\mathbf{v}_j^T\mathbf{W}_k^T\mathbf{u}_i + 2r_{jk}\mathbf{u}_i^T\mathbf{W}_k\mathbf{v}_j + r_{jk}^2
    \\&
    -{\zeta^k_{ij}}^2)+\log\sigma (\zeta^k_{ij})).
    \end{split}
\end{equation}
Finally, by plugging $\log\Tilde{p}(\bm{\mathcal{D}|\theta})$ into Eq. \ref{eq:vb-update-rule}, we receive the new \emph{bounded} version of the VB update rule:
\begin{equation}
\label{eq:vb-bounded-update-rule}
\tilde{q}^*(z)=\exp (\mathbb{E}_{q(\bm{\theta}\setminus \{z\})}[\log \Tilde{p}(\bm{\mathcal{D}}|\bm{\theta}) + \log p(\bm{\theta}|\mathcal{H})] + \text{const}).
\end{equation}

\subsubsection{\textbf{Optimization}}
\label{subsec:optimization}
\hfill\\
The update rule in Eq.~\ref{eq:vb-bounded-update-rule} enables conjugacy, and hence we are able to recognize the natural parameters of the Normal distributions and the Gamma distributions that give raise to the update steps for each of the approximated posterior distributions $q(z), \,  z\in\bm{\theta}$:

\textbf{\underline{Updating $q(\mathbf{u}_i)$ and $q(\mathbf{v}_j)$}}:
The updates of $q(\mathbf{u}_i)$ are based on the sufficient statistics of the multivariate Gaussian, namely the precision matrix $\mathbf{P}_{\mathbf{u}_i}$ and mean vector $\bm{\mu}_{\mathbf{u}_i}$ given by:
\begin{equation}
\label{eq:ui-update}
\begin{split}
\mathbf{P}_{\mathbf{u}_i} &= \mu_{\tau_{u_i}}\mathbf{I} + 2(\textstyle\sum_{j:(i,j)\in I_D} \lambda(\xi_{ij}) \mathbb{E}[\mathbf{v}_j\mathbf{v}_j^T]
\\&
\qquad\qquad\quad+ \textstyle\sum_{k=1}^{N_R} \textstyle\sum_{j:(i,j)\in I_G^k} \lambda(\zeta^k_{ij})\E[\mathbf{W}_k\mathbf{v}_j\mathbf{v}_j^T\mathbf{W}_k^T]),
\\
\bm{\mu}_{\mathbf{u}_i}&= \mathbf{P}^{-1}_{\mathbf{u}_i}(|\pi_i|^{-1}\mu_{\tau_{u_i}}\textstyle\sum_{m\in\pi_i}\bm{\mu}_{\mathbf{h}_m^u} 
\\&
\quad\quad\quad+\half(\textstyle\sum_{j:(i,j)\in I_D}(d_{ij}-4\lambda(\xi_{ij})\mu_{b_j})\bm{\mu}_{\mathbf{v}_j} 
\\&
\quad\quad\quad\quad\quad+\textstyle\sum_{k=1}^{N_R} \textstyle\sum_{j:(i,j)\in I_G^k}(g^k_{ij}  -4\lambda(\zeta^k_{ij})\mu_{r_{jk}})\mathbf{M}_k\bm{\mu}_{\mathbf{v}_j}),
\end{split}
\end{equation}
with
\begin{equation}
\label{eq:wvvw-expectation}
\begin{split}
\E[\mathbf{W}_k\mathbf{v}_j\mathbf{v}_j^T\mathbf{W}_k^T]&=\E[\textstyle\sum_{m=1}^{t_k}\mathbf{x}_{mk}\mathbf{y}_{mk}^T\mathbf{v}_j\mathbf{v}_j^T\mathbf{y}_{mk}\mathbf{x}_{mk}^T
\\&\quad\quad
+\textstyle\sum_{m\neq n}\mathbf{x}_{mk}\mathbf{y}_{nk}^T\mathbf{v}_j\mathbf{v}_j^T\mathbf{y}_{nk}\mathbf{x}_{mk}^T]
\\&
=\textstyle\sum_{m=1}^{t_k}\E[(\mathbf{y}_{mk}^T\mathbf{v}_j)^2]\E[\mathbf{x}_{mk}\mathbf{x}_{mk}^T]
\\&\quad
+\textstyle\sum_{m \neq n} \bm{\mu}_{\mathbf{x}_{mk}}\bm{\mu}_{\mathbf{y}_{mk}}^T\E[\mathbf{v}_j\mathbf{v}_j^T]\bm{\mu}_{\mathbf{y}_{nk}}\bm{\mu}_{\mathbf{x}_{nk}}^T,
\end{split}
\end{equation}
and $\mathbf{M}_k=\mathbf{M}_{\mathbf{X}_k}\mathbf{M}_{\mathbf{Y}_k}^T$, where $\mathbf{M}_{\mathbf{X}_k}$ and $\mathbf{M}_{\mathbf{Y}_k}$ are the matrices whose columns are the expectations $\bm{\mu}_{\mathbf{x}_{mk}}$ and $\bm{\mu}_{\mathbf{y}_{mk}}$. 

In practice, we only consider diagonal precision matrices by setting all off-diagonal parameters to zero\footnote{In our experiments, we did not observe any degradation while using diagonal precision matrices instead of full precision matrices.}. Besides reducing the model's space complexity, it also simplifies significantly the matrix inversion for extracting the mean value. %
The update for $q(\mathbf{v}_j)$ is symmetric to that of $q(\mathbf{u}_i)$ (as appears in Eq.~\ref{eq:ui-update}).

\textbf{\underline{Updating $q(\mathbf{x}_{mk})$ and $q(\mathbf{y}_{mk})$}}: The update of $q(\mathbf{x}_{mk})$ is given by: 
\begin{equation}
\label{eq:x-update}
\begin{split}
\mathbf{P}_{\mathbf{x}_{mk}} &= \mu_{\tau_{x_{mk}}}\mathbf{I} +  2\textstyle\sum_{(i,j)\in I_G^k} \lambda(\zeta^k_{ij})\E[\mathbf{u}_i\mathbf{y}_{mk}^T\mathbf{v}_j\mathbf{v}_j^T\mathbf{y}_{mk}\mathbf{u}_i^T],
\\
\bm{\mu}_{\mathbf{x}_{mk}}&= \frac{1}{2}\mathbf{P}^{-1}_{\mathbf{x}_{mk}}(
\textstyle\sum_{(i,j)\in I_G^k}g^k_{ij}\bm{\mu}_{\mathbf{y}_{mk}}^T\bm{\mu}_{\mathbf{v}_j}\bm{\mu}_{\mathbf{u}_i}^T
\\&\qquad\qquad\qquad\quad\quad
-4\lambda(\zeta^k_{ij})(
\mu_{r_{jk}}\bm{\mu}_{\mathbf{y}_{mk}}^T\bm{\mu}_{\mathbf{v}_j}\bm{\mu}_{\mathbf{u}_i}^T
\\&\qquad\qquad\qquad\qquad\qquad\quad
+ 
\sum_{n\neq m}\bm{\mu}_{\mathbf{x}_{nk}}^T\E[\mathbf{u}_i\mathbf{v}_j^T\bm{\mu}_{\mathbf{y}_{nk}}\bm{\mu}_{\mathbf{y}_{mk}}^T\mathbf{v}_j\mathbf{u}_i^T]
),
\end{split}
\end{equation}
where the expectations are computed in the same manner as in Eq.~\ref{eq:wvvw-expectation}. The update of $q(\mathbf{y}_{mk})$ is symmetric.

\textbf{\underline{Updating $\xi_{ij}$ and $\zeta^k_{ij}$}}:
The updates of the variational parameters are given by:
\begin{equation}
\label{eq:ui-update2}
\begin{split}
\xi_{ij}&=(\E[(\mathbf{u}_i^T\mathbf{v}_j)^2]
+ 2\mu_{b_j}\bm{\mu}_{\mathbf{u}_i}^T\bm{\mu}_{\mathbf{v}_j} + \E[b_j^2])^{1/2},
\\
\zeta^k_{ij}&=(\E[(\mathbf{u}_i^T\mathbf{W}_k\mathbf{v}_j)^2] + 2\mu_{r_{jk}}\bm{\mu}_{\mathbf{u}_i}^T\mathbf{M}_k\bm{\mu}_{\mathbf{v}_j}
+ \E[r_{jk}^2])^{1/2}. 
\end{split}
\end{equation}

\textbf{\underline{Updating $q(\mathbf{h}^u_i)$ and $q(\mathbf{h}^v_i)$}}:
The updates of $q(\mathbf{h}^u_i)$ and $q(\mathbf{h}^v_i)$ are symmetric as well. Hence, we only provide the update of $q(\mathbf{h}^u_i)$ based on the precision matrix $\mathbf{P}_{\mathbf{h}^u_i}$ and mean vector $\bm{\mu}_{\mathbf{h}^u_i}$:
\begin{equation}
\label{eq:hu-updates}
\begin{split}
\mathbf{P}_{\mathbf{h}^u_i}&= (\mu_{\tau_{h^u_i}}+\textstyle\sum_{m\in\omega_i}(\mu_{\tau_{h^u_m}}+\mu_{\tau_{u_m}})|\pi_m|^{-2})\mathbf{I},  \\
\bm{\mu}_{\mathbf{h}^u_i}&=\mathbf{P}^{-1}_{\mathbf{h}^u_i}(|\pi_i|^{-1}\mu_{\tau_{h^u_i}}\textstyle\sum_{l\in\pi_i}\bm{\mu}_{\mathbf{h}_l^u}
\\&\qquad\quad
+\textstyle\sum_{m\in\omega_i}|\pi_m|^{-1}(\mu_{\tau_{u_m}}\bm{\mu}_{\mathbf{u}_m}+\mu_{\tau_{h^u_m}}\bm{\mu}_{\mathbf{h}^u_m})
\\&\qquad\qquad\qquad\quad
-|\pi_m|^{-2}(\mu_{\tau_{h^u_m}}+\mu_{\tau_{u_m}})\textstyle\sum_{n\in\pi_m\setminus\{i\}}\bm{\mu}_{\mathbf{h}^u_n}).
\end{split}
\end{equation}

\textbf{\underline{Updating $q(b_j)$}}:
The update of $q(b_j)$ is given by the (scalar) precision $\mathrm{p}_{b_j}$ and mean $\mu_{b_j}$:
\begin{equation}
\label{eq:bj-updates}
\begin{split}
\mathrm{p}_{b_j}&=\mu_{\tau_{b_j}} + 2\textstyle\sum_{i:(i,j)\in I_D}\lambda(\xi_{ij}),
\\
\mu_{b_j}&=\half\mathrm{p}^{-1}_{b_j}\textstyle\sum_{i:(i,j)\in I_D}(d_{ij}-4\lambda(\xi_{ij})\bm{\mu}^T_{\mathbf{u}_i}\bm{\mu}_{\mathbf{v}_j})
\end{split}
\end{equation}

\textbf{\underline{Updating $q(r_{jk})$}}:
The update of $q(r_{jk})$ is given by the scalar parameters:
\begin{equation}
\label{eq:r_jk-updates}
\begin{split}
\mathrm{p}_{r_{jk}}&=\mu_{\tau_{r_{jk}}} + 2\textstyle\sum_{i:(i,j)\in I_G^k}\lambda(\zeta_{ij}^k),
\\
\mu_{r_{jk}}&=\half\mathrm{p}^{-1}_{b_j}\textstyle\sum_{i:(i,j)\in I_G^k}(g_{ij}^k-4\lambda(\zeta_{ij}^k)\bm{\mu}^T_{\mathbf{u}_i}\mathbf{M}_k\bm{\mu}_{\mathbf{v}_j})
\end{split}
\end{equation}

\textbf{\underline{Updating $q(\tau_{u_i}),q(\tau_{v_i}),q(\tau_{h^u_i})$, and $q(\tau_{h^v_i})$}}:
$q(\tau_{u_i})$ follows the Gamma distribution with rate $\alpha_{u_i}$ and shape $\beta_{u_i}$ parameters. Their updates are given by
\begin{equation}
\label{eq:tau_u-updates}
\begin{split}
&\alpha_{u_i}=\alpha+\half t,\\& \beta_{u_i}=\beta+\half\E[(\mathbf{u}_i-\mathbf{s}^u_i)^T(\mathbf{u}_i-\mathbf{s}^u_i)].
\end{split}
\end{equation}
The updates of $q(\tau_{v_i}),q(\tau_{h^u_i})$, and $q(\tau_{h^v_i})$ are symmetric to $q(\tau_{u_i})$.

\textbf{\underline{Updating  and $q(\tau_{x_{mk}})$ and $q(\tau_{y_{mk}})$}}:
The updates of $q(\tau_{x_{mk}})$ are given by:
\begin{equation}
\label{eq:tau_x-updates}
\begin{split}
\alpha_{x_{mk}}&=\alpha+\half t,\\ \beta_{x_{mk}}&=\beta+\half\E[\mathbf{x}_{mk}^T\mathbf{x}_{mk}].
\end{split}
\end{equation}
The updates of $q(\tau_{y_{mk}})$ are symmetric.

\textbf{\underline{Updating $q(\tau_{b_j})$ and $q(\tau_{r_{jk}})$}}:
The updates of $q(\tau_{b_j})$ are given by:
\begin{equation}
\label{eq:tau_b-updates}
\begin{split}
\alpha_{b_j}&=\alpha+\half,
\\
\beta_{b_j}&=\beta+\half \E[b^2_j].
\end{split}
\end{equation}
The updates of $q(\tau_{r_{jk}})$ are symmetric.

\vspace{3mm}
VBN's optimization algorithm alternates between the '$u$' and '$v$' parts of the VBN graph. The optimization easily scales as the updates of the $u$ ($v$) part are embarrassingly parallel in the number of entities $N_W$ and relations $N_R$. Specifically, we can update each of the $q$ distributions in $\mathbf{U},\mathbf{V}, \mathbf{B},\mathbf{R},\bm{\Tau}^u,\bm{\Tau}^v,\bm{\Tau}^b,\bm{\Tau}^r,\bm{\Tau}^{h^u},\bm{\Tau}^{h^v}$, each in turn, in parallel, by using the expected value of the variables in the other sets (including $\mathbf{H}^u,\mathbf{H}^v,\mathbf{X},\mathbf{Y}$). Yet, the updates of the variables in $\mathbf{H}^u,\mathbf{H}^v,\mathbf{X},\mathbf{Y}$ require special care: This is since according to Eq.~\ref{eq:hu-updates}, the updates of a variable in $\mathbf{H}^u$ ($\mathbf{H}^v$) depend on the expectations of its children, parents, and other parents of its children (if exist).

In order to enable parallel updates of $\mathbf{H}^u$, we propose Alg.~\ref{alg:partition} that operates in a bottom-up manner and partitions $A^u = \mathbf{H}^u \cup \mathbf{U}$ to $M$ disjoint sets $\{Y^u_m\}_{m=1}^M$ s.t. for each variable $n$ in the set $Y^u_m$, its parents (line 20), children, and other parents of its children (line 15) are \emph{not} in $Y^u_m$. This ensures all the variables in $Y^u_m$ can be updated in parallel. Note that $\mathbf{U}\subset Y^u_1$ is guaranteed, hence, $\mathbf{U}$ can be updated in parallel. Then, Alg.~\ref{alg:partition} is applied to $A^v=\mathbf{H}^v \cup \mathbf{V}$ to produce $\{Y^v_m\}_{m=1}^M$. In practice, we perform Alg.~\ref{alg:partition} once before the optimization procedure begins and save the indices of the nodes for each disjoint set for both the $u$ and $v$ parts of the graph. %
In addition, as can be seen in Eq.~\ref{eq:x-update}, $\bm{\mu}_{\mathbf{x}_{mk}}$ depends on $\bm{\mu}_{\mathbf{x}_{nk}}$, for all $n\neq m$. Therefore, parallel updates of the variables in $\mathbf{X}_k$ ($\mathbf{Y}_k$) are possible across the $k$ axis, but not across $m$.
\def\NoNumber#1{{\def\alglinenumber##1{}\State #1}\addtocounter{ALG@line}{-1}}
\renewcommand{\algorithmicrequire}{\textbf{Input:}}
\renewcommand{\algorithmicensure}{\textbf{Output:}}

\begin{algorithm}[t]
  \caption{Partition Algorithm}
  \begin{algorithmic}[1]
    \Require $A$ - set of variables in a graph
    \Ensure $\{Y_m\}_{m=1}^M$ - A partition of $A$ to $M$ distinct sets
    \State $M \gets 1$, $W\gets\emptyset$, $S_{r}\gets\emptyset$
    \State$S \gets $all leaf variables in $A$
    \While{$S\neq\emptyset$}
        \State $Y_M \gets S$
        \State $W\gets W \cup S$
        \State $S \gets S_{r}$
        \State $S_{r}\gets\emptyset$, $S_{c}\gets\emptyset$ 
        \For{$n$ in $Y_M$}
        \State \parbox[t]{200pt}{\raggedright{$S\gets S\cup\mathcal{P}(n)$ \qquad\qquad\quad\color{blue}\#$\mathcal{P}(n)$ are the parents of $n$}}
        \EndFor
        \For{$n$ in $S$}
            \If {$n \in W$}
                \State $S\gets S\setminus \{n\}$
            \ElsIf {$\mathcal{C}(n) \cap S_{c} \neq \emptyset$} \color{blue}\# $\mathcal{C}(n)$ are the children of $n$\color{black}
                \State $S\gets S\setminus \{n\}$
                \State $S_r\gets S_r\cup\{n\}$
            \Else
                \State $S_{c} \gets S_{c} \cup \mathcal{C}(n)$  
            \EndIf
            \State $S\gets S \setminus \mathcal{P}(n)$
        \EndFor
        \State $M \gets M+1$
        \EndWhile
        \State \Return $\{Y_m\}_{m=1}^M$

\end{algorithmic}
  \label{alg:partition}
\end{algorithm}

Finally, the VBN optimization algorithm is presented in Alg.~\ref{alg:opt}. The algorithm receives the partitions $\{Y^u_m\}_{m=1}^M$ and $\{Y^v_m\}_{m=1}^M$, the Gamma hyperprior parameters $\alpha$ and $\beta$ (that are set to 1 in our implementation), and the number of epochs $T$. First, the parameters for each $q(z),z\in\bm{\theta}$ are initialized. Specifically, for normal variables, we sample the mean from the standard multivariate normal distribution and set the precision to the identity matrix. Both the shape and rate parameters of the Gamma variables were initialized to 1.
At every epoch, each $q(z),z\in\bm{\theta}$ is being updated once (the computation depends on the dependencies that are dictated by in the update equations). The partition to independent sets (Alg.~\ref{alg:partition}), allows easy parallelism of the updates of each group of variables. Note that the variational parameters $\xi_{ij}$ and $\zeta_{ij}^k$ are computed on-the-fly when needed.
In our experiments, Alg.~\ref{alg:opt} converged after 30-40 epochs, depending on the dataset.  

\begin{algorithm}[t]
  \caption{VBN Optimization}
  \begin{algorithmic}[1]
     \Require $\{Y^u_m\}_{m=1}^M, \{Y^v_m\}_{m=1}^M$, $\alpha, \beta, T$

    \State Initialize all the normal variables with an identity precision, and mean that is sampled from the standard normal distribution.
    \State Initialize Gamma variables with shape and rate parameters equal to 1.
    \For{$n=1:T$}
    \State \parbox[t]{200pt}{\raggedright{Sample $\bm{\mathcal{D}}$ according to the description from Secs.~\ref{subsec:cooccurrence} and \ref{subsec:explicit}}}
    \For{$m=1:M$} \quad\# $q(\mathbf{u}_i),q(\mathbf{h}^u_i)$ updates
    \State \parbox[t]{200pt}{\raggedright{Update all $q(y),y\in Y^u_m$, using Eqs.~\ref{eq:ui-update2} and \ref{eq:hu-updates},\\ in parallel}}
    \EndFor
    \State \parbox[t]{200pt}{\raggedright{Update all $q(\mathbf{v}_i),q(\mathbf{h}^v_i)$ by performing a symmetric version of Steps 5-7}}
    \For{$m=1:\max(\{t_k\})$}
    \State \parbox[t]{200pt}{\raggedright{Update all $q(\mathbf{x}_{mk}),k\in \{1,...,N_R\}$, using Eq.~\ref{eq:x-update},\\ in parallel}}
    \EndFor 
    \State \parbox[t]{200pt}{\raggedright{Update all $q(\mathbf{y}_{mk})$ by performing a symmetric version of Steps 9-11}} 
    \State \parbox[t]{200pt}{\raggedright{Update all $q(b_j)$ and $q(\mathbf{r}_{jk})$, in parallel, using Eqs.~\ref{eq:bj-updates} and \ref{eq:r_jk-updates}, respectively}}
    \State \parbox[t]{200pt}{\raggedright{Update all $q(\tau_{u_i})$, $q(\tau_{h^u_i})$, $q(\tau_{v_i})$, $q(\tau_{h^v_i})$, $q(\tau_{x_{mk}})$, $q(\tau_{y_{mk}})$ $q(\tau_{b_j})$, $q(\tau_{r_{jk}})$, in parallel, using Eqs.~\ref{eq:tau_u-updates}, \ref{eq:tau_x-updates}, and \ref{eq:tau_b-updates} (and their symmetric versions)}}

    \EndFor

\end{algorithmic}
  \label{alg:opt}
\end{algorithm}

\subsubsection{\textbf{Approximate Bayesian Inference}}
\label{subsec:inference}

\hfill\\
Once Alg.~\ref{alg:opt} converged, we approximate the posterior predictive integral from Eq.~\ref{eq:posterior-predicitive} by:
\begin{equation}
\label{eq:approximated-posterior-predicitive}
\begin{split}
    p(d_{ij}=1|\bm{\mathcal{D}}, \mathcal{H})&\underset{(1)} {\approx}  \int{ \sigma(\mathbf{u}_i^T\mathbf{v}_j+b_j)q(\mathbf{u}_i)q(\mathbf{v}_j)q(b_j)d\mathbf{u}_id\mathbf{v}_jdb_j} \\ & \underset{(2)}{\approx}  \int{ \sigma(x)\mathcal{N}(x;\mu_x,\sigma^2_x)dx}  \underset{(3)}{\approx}   \sigma(\mu_x/\sqrt{1+\pi\sigma^2_x/8}).
\end{split}
\end{equation}
Equation~\ref{eq:approximated-posterior-predicitive} presents three approximations: approximation (1) is the VB approximation that replaces $p(\bm{\theta}|\bm{\mathcal{D}},\mathcal{H})$ with the fully factorized $q(\bm{\theta})$. Hence, each $z\in\bm{\theta}\setminus\{\mathbf{u}_i,\mathbf{v}_j,b_j\}$ is integrated out. 
Yet, we are still left with an intractable integral. Therefore, we propose to approximate the distribution of $x=\mathbf{u}_i^T\mathbf{v}_j+b_j$ with a normal distribution (based on the first and second moments), which leads to approximation (2). The theoretical justification for this step is due to the Berry-Esseen Theorem \cite{berry1941accuracy} that places a bound on the error of a Gaussian approximation to the sum of independent variables. This bound is inversely proportional to the number of summands. In our case, $\mathbf{u}_i^T\mathbf{v}_j+b_j$ is a sum of $t+1$ independent variables. Hence, as the dimension $t$ increases, the error induced by approximation (2) decreases (we refer the reader to Theorem~\ref{th:berry-theorem} for further details). Finally, approximation (3) follows from \cite{mackay1992evidence}.

Next, we justify approximation (2) in Eq: \ref{eq:approximated-posterior-predicitive}. 
\begin{theorem}
\label{th:berry-theorem}
\text{\cite{berry1941accuracy}} Let $x_1,...,x_n$ be independent random variables with $\mathbb{E}[x_i]=0, \mathbb{E}[x_i^2]=\sigma_i^2>0$, and $\mathbb{E}[|x_i|^3]=\rho_i<\infty$. Also, let $S_n=\frac{\sum_{i=1}^{n}x_i}{\sqrt{\sum_{i=1}^{n}\sigma_i^2}}$. Denote $F_n$ and $\Phi$ the CDFs of $S_n$ and of the standard normal distribution, respectively. Then, 
\begin{equation*}
    \exists C>0 \; \forall n\in\mathbb{N}: \sup_{x\in\mathbb{R}} |F_n(x)-\Phi(x)|\leq \frac{MC}{\sqrt{\sum_{i=1}^{n}\sigma_i^2}},
\end{equation*}
where $M=\max_{1\leq i \leq n} \frac{\rho_i}{\sigma_i^2}$.
\end{theorem}
Therefore, as the number of summands $n$ increases, the maximal difference between $F_n$ and the standard normal CDF decreases. 
Denote
$
    x=\mathbf{u}_i^T\mathbf{v}_j+b_j=b_j + \sum_{k=1}^t u_{ik}v_{jk}.
$
While $b_j$ is a normal variable, each term $u_{ik}v_{jk}$ in the summation is a product of normal variables and hence does not follow the normal distribution. Yet, this summation considers independent\footnote{As explained in Section \ref{subsec:approximate bayesian inference}, we consider diagonal precision matrices.} random variables that meet the conditions in Theorem \ref{th:berry-theorem}\footnote{While the inequality in Theorem \ref{th:berry-theorem} is stated with standardized variables, the bound on the absolute difference is not affected by a scaling factor nor by a shift.}. Therefore, we approximate the density of $x$ with the normal PDF (according to its first and second moments). Empirically, we found that the histogram induced by a Monte-Carlo sampling from the true PDF of $x$ matches very closely its normal approximation for all $t>20$ (in our experiments we used $t=50$).

\section{Experimental Setup}
\label{sec:experiments}
In this section we describe the datasets, evaluation tasks, evaluated models, and hyperparameters configuration.

\subsection{Datasets}
\label{subsec:datasets}
Our evaluation covers NLU, recommender systems, and medical informatics datasets as follows: 

{\bf \underline{NLU:}}  
The SemCor dataset~\cite{miller1993semantic} consists of 37,176 annotated sentences, with 820,411 words and a vocabulary size of 43,416. The hierarchical and explicit relations were extracted from WordNet~\cite{miller1990introduction}.%

{\bf \underline{Recommender Systems:}}
Two popular collaborative filtering datasets were used: MovieLens 25M~\cite{harper2015movielens} and Yahoo! Music~\cite{yahoo2008music}, consisting of user-item ratings on a scale of [0-5] and [0-100] for movies and songs, respectively. For each dataset, we sampled a set of 4,000 items and 6,000 users. For every user, we considered movies (songs) that were ranked above 3.5 (80) as items that co-occur together. The movies and songs datasets further provide hierarchical information in the form of \textit{genres $\rightarrow$ director $\rightarrow$ movie} and \textit{artist $\rightarrow$ album $\rightarrow$ song}, respectively.

{\bf \underline{Medical Informatics:}} The MIMIC-III dataset~\cite{johnson2016mimic} is based on patients admitted to intensive care units at a large tertiary care hospital. It is comprised of 13,000 diagnoses made for 46,520 patients. The diagnoses are labeled by ICD9\footnote{\url{www.cdc.gov/nchs/icd/icd9.htm}} codes and the ICD9 classification index provides hierarchical information for these diagnoses.

\subsection{Evaluated Models and Hyperparameter Configuration}
\label{baselines}
For every dataset, we defined a unique held-out validation-set that was used to optimize the hyperparameters of each model. Our evaluation include the following models:
\begin{itemize}
    \item {\bf VBN:} Our model with a two-level hierarchy e.g., \textit{animal $\rightarrow$ dog $\rightarrow$ Poodle} (our experiments using single- or three-level hierarchies did not yield improvements). In the Yahoo! Music and the MovieLens datasets, the hierarchical information are as explained in Sec.~\ref{subsec:datasets}.
In the MIMIC-III medical dataset, we employed the ICD9 hierarchy over the entities taken from Wikipedia\footnote{\url{en.wikipedia.org/wiki/List_of_ICD-9_codes}}. 
Unlike the NLU dataset, these datasets do not include explicit relations between entities. For the NLU dataset, hierarchical relations are based on \textit{hypernyms} and \textit{troponyms}, extracted from WordNet~\cite{miller1995wordnet}. Explicit relations between entities had been extracted from WordNet as well. 
Specifically, the lexical-semantic relations utilized in our experiments were \textit{antonymy} and \textit{meronymy}.
\item {\bf SG and SG-R:}
The Skip-Gram with negative sampling method from \cite{mikolov2013distributed}, and its Retrofitted version (SG-R). As explained in Section~\ref{sec:related}, retrofitting~\cite{faruqui-etal-2015-retrofitting} enables the incorporation of side information as a post-processing step, providing an alternative to VBN's mechanism for modeling hierarchical side-information.
\item {\bf BSG and BSG-R:}
The Bayesian Skip-Gram model \cite{barkan2017bayesian}, and its Retrofitted version (BSG-R).
\end{itemize}

For all datasets and models, we found out that the optimal hyperparameter configuration is an embedding size of $t=50$, negative-to-positive ratio of 1, and Gamma hyperprior parameters with $\alpha=\beta=1$. In our experiments, all models converged after 40-80 epochs, depending on the dataset characteristics (number of entities, number of interactions, etc.). Following convergence, we retrofitted the embeddings produced by SG and BSG using the post-processing procedure from~\cite{faruqui-etal-2015-retrofitting} to produce SG-R and BSG-R.

\subsection{Evaluation Tasks and Measures}\label{sec:tasks}
We evaluated the quality of the learned representations across various tasks. For all the benchmarks, we considered 80\%/20\% train/test split. Then, 15\% of the train data was kept aside as validation data while the rest was used for training the model. Namely, the effective split consists of 68\%/12\%/20\% train/validation/test.
In what follows, we describe the evaluation tasks.

\subsubsection{\bf Inference Tasks}
The models are evaluated on three inference tasks:
\begin{itemize}
    \item {\bf Sentence Completion:} 
In this task, the goal is to complete a masked word in a sentence. The test sentences are taken from the NLU dataset.
\item {\bf Recommendations:}
We trained the models on a masked version of the datasets, where for each user two items were masked: The first masked item was randomly sampled from the user's list and the second masked item was the last item consumed by the user. Then, the task was to recommend the masked items for each user.
\item {\bf Medical Inference:}
We masked the patients' diagnoses in the same manner as done for the recommendation task. Similarly, the task was to predict the correct (masked) diagnoses for each patient.
\end{itemize}
For inference tasks, the queries are the masked sequences. The sentences, users, and patients were represented by the average of their words, items, or diagnoses vectors (for SG) and random variables (for BSG and VBN). The score between a query and a candidate is computed by the cosine similarity for SG and by Eq.~\ref{eq:approximated-posterior-predicitive} for BSG and VBN.

The models' performance is evaluated according to the following measures:
\begin{itemize}
    \item {\bf Hit-Rate at \emph{k}\% (HR@\emph{k}\%):} This measure outputs $1$ if the correct item to be retrieved is ranked in the top $k\%$ percentile, otherwise $0$. Then, the average HR@$k\%$ score is taken across the test set examples.
    \item {\bf Mean Percentile Rank (MPR):}
This measure outputs $1$ minus the percentile rank (PR) of the test example. The PR is the rank of the hidden item divided by the catalog size. The MPR score is the average of the PR scores for all the test set examples.
\end{itemize}

\subsubsection{\bf Word Similarity Tasks}
We further evaluated the models on five word similarity datasets: WordSim-353 (WS)~\cite{finkelstein2001placing}, Stanford's Contextual Word Similarities (SCWS)~\cite{huang2012improving}, Rare Words (RW)~\cite{luong2013better}, MEN~\cite{bruni2014multimodal} and SimLex-999 (SL)~\cite{hill2015simlex}. Each dataset contains word pairs that are associated with human annotated similarity scores, considered as ground truth. The model's performance is evaluated based on the Spearman's correlation between the ground truth scores and the scores produced by the models. 

\section{Results}
\label{sec:results}
In this section we provide the experimental results as well as a qualitative analysis for assessing the performance advantages of our VBN model.
\definecolor{mygreen}{rgb}{0.0, 0.4, 0.0}
\begin{table}[t]
 \setlength{\tabcolsep}{0.7pt}
\caption{HR@10\% and MPR results for inference tasks.}\vspace{-3mm}
\label{hitrate-table}
\centering
\large
\begin{tabular}{lcccc|cccc}
\toprule 
 & \textsc{Movie} & \textsc{Music} & \textsc{Medic} & \textsc{Text} & \textsc{Movie} & \textsc{Music} & \textsc{Medic} & \textsc{Text}\\
\midrule
\multicolumn{9}{c}{\text{\quad\quad\quad\qquad \underline{HR@10}\%\quad\quad\quad}}\\
\multicolumn{9}{c}{\text{\quad\quad\quad--- Full catalog --- \qquad \quad\quad\quad--- Rare catalog ---}}\\
\textsc{VBN}    & \textbf{39.4} & \textbf{63.5} & \textbf{42.2} & \textbf{39.2}  & \textbf{25.1} & \textbf{45.7} & \textbf{28.3}  &  \textbf{30.8}\\
\textsc{BSG}   & 33.0 & 57.1 & 37.5 & 37.6& 13.7 & 31.6 & 17.3 & 20.1 \\
\textsc{BSG-R}   & 33.2 & 57.6 & 37.9 & 37.6& 13.8 & 32.1 & 17.4 & 20.6 \\
\textsc{SG}    & 22.7 & 46.1 & 23.3 &  23.6& 8.0 & 29.6 & 15.6 &20.0\\
\textsc{SG-R}    & 22.8 & 47.2 & 23.7 &  23.8& 8.4 & 29.9 & 15.7 &20.8\\
\midrule
\midrule
\multicolumn{9}{c}{\text{\quad\quad\quad\qquad \underline{MPR}\quad\quad\quad}}\\
\multicolumn{9}{c}{\text{\quad\quad\quad--- Full catalog --- \qquad \quad\quad\quad--- Rare catalog ---}}\\
\textsc{VBN}  & \textbf{79.6} & \textbf{77.9} & \textbf{73.9} & \textbf{74.2}& \textbf{64.7} & \textbf{64.5} & \textbf{68.5}  & \textbf{64.4}\\
\textsc{BSG} & 77.1 & 73.4 & 68.9 & 71.1& 61.2 & 57.0 & 62.0 & 60.1 \\
\textsc{BSG-R} & 77.9 & 74.2 & 69.4 & 71.9& 62.1 & 57.8 & 62.9 & 60.9 \\
\textsc{SG}  & 75.5 & 58.3 & 68.8 &  64.2& 54.1 & 53.9 & 56.4 & 52.1\\
\textsc{SG-R}  & 75.7 & 58.5 & 68.9 &  64.5& 59.6 & 55.3 & 57.6 & 54.7\\

\bottomrule

\end{tabular}
\end{table}

\subsection{Inference Tasks}
\label{subsec:infr}
Table~\ref{hitrate-table} presents the HR@$10\%$, and MPR values, for each combination of a model and a dataset. We consider two test sets: the original test set, `Full', and the `Rare' dataset, comprising the 20\% least frequent entities.
We can observe that VBN outperforms all baselines across all datasets. This showcases the importance of 1) Modeling external taxonomy information: VBN outperforms all baselines that do not utilize taxonomy (BSG and SG) and 2) Bayesian treatment: VBN outperforms SG and SG-R. In the same manner, BSG improves upon SG which demonstrates the merit of the Bayesian approach over simple point estimate solutions (BSG $\succ$ SG). Finally, VBN outperforms the other methods over long-tail entities (the 'Rare' catalog), where the gap between VBN and the other baselines becomes even more significant. These results demonstrate the ability of VBN to provide better modeling for entities in the long-tail. 

\subsection{Word Similarity}

\begin{table}[t]
 \setlength{\tabcolsep}{1.pt}
 \caption{Word similarity evaluation results.}\vspace{-3mm}
\label{wordsim-table}
\begin{center}

\large
\begin{tabular}{lccccc|r}
\toprule
\textsc{Model} & \textsc{MEN} & \textsc{RW} & \textsc{SCWS} & \textsc{SL} & \textsc{WS} & \textsc{AVG}\\
\midrule
\multicolumn{6}{c}{\text{\qquad\qquad--- Full catalog --- }}\\
\textsc{VBN}    & \textbf{44.7} & \textbf{30.2} & \textbf{43.2} & \textbf{15.6} & \textbf{38.2} & \textbf{34.4}\\
\textsc{BSG} & 38.4 & 27.6 & 39.6 & 14.2 & 28.6 & 29.7  \\
\textsc{BSG-R} & 39.2 & 27.2 & 41.9 & 14.6 & 31.5 & 30.9 \\
\textsc{SG}    & 38.3 & 23.0 & 36.2 & 13.6 & 27.0 & 27.6  \\
\textsc{SG-R}    & 38.7 & 27.7 & 39.0 & 13.9 & 28.4 & 29.5  \\
\midrule
\midrule
\multicolumn{6}{c}{\text{\qquad\qquad--- Rare catalog --- }}\\
\textsc{VBN}    & \textbf{37.7} & \textbf{27.3} & \textbf{31.7} & \textbf{12.9} & \textbf{42.9} & \textbf{30.5} \\
\textsc{BSG} & 27.9 & 24.9 & 27.4 & 11.0 & 32.4 & 24.7\\
\textsc{BSG-R}& 28.2 & 25.3 & 28.5 & 11.8 & 34.0 & 25.6\\
\textsc{SG}   & 27.2 & 24.2 & 23.1 & 9.2 & 22.5 & 21.2 \\
\textsc{SG-R}    & 27.5 & 24.4 & 25.6 & 10.5 & 30.8 & 23.8\\

\bottomrule
\vspace{-4mm}
\end{tabular}
\end{center}
\end{table}

Table \ref{wordsim-table} presents the word similarity results\footnote{These results are sub-optimal compared to those reported in previous word embedding works (e.g., \cite{pennington2014glove}) due to the use of a significantly smaller corpus.}. The last column depicts the average across all datasets. The best performing results are boldfaced, and evidently, VBN outperforms other methods across all datasets.
The retrofitting stage improves both SG and BSG, which indicates the effectiveness of incorporating lexical knowledge into the word representations.
Nonetheless, BSG surpasses SG with or without retrofitting, which demonstrates the merits of Bayesian modeling. 
Importantly, VBN outperforms both BSG-R and SG-R indicating the benefit of learning relational knowledge during training rather than as a post-processing step. 
Finally, we can observe that in the case of rare words, the gap between VBN and the other models' performances increases.

\subsection{Learning Explicit Relations}
\label{subsec:rel}
Recall that VBN explicitly models relations of entities. We evaluate this capability by extracting \emph{opposites} (\emph{antonyms}) pairs from WordNet.
We split the antonyms in the Brown dataset to 3,279 and 1,094 train and test pairs, respectively. 
Given a query word $i$, we rank a candidate word $j$ by $p(g^{opp}_{ij}=1|\bm{\mathcal{D}},\mathcal{H})$, where the predictive probabilities are computed using the application of the approximation from Eq.~\ref{eq:approximated-posterior-predicitive} to $p(g^{opp}_{ij}=1|\bm{\mathcal{D}},\mathcal{H})$.

In order to quantify the contribution from modeling explicit relations, we further consider an ablated version of VBN without the explicit relations component. To this end, we simply omit the terms that account for the explicit relations from the joint log-likelihood. Then, we rank a candidate word $j$ by $p(d_{ij}=1|\bm{\mathcal{D}},\mathcal{H})$.

Table~\ref{hitrate-relations-table} presents the MPR and HR@\emph{k\%} results. As expected, incorporating the information from explicit relations during training (VBN) leads to significantly better w.r.t the ablated version of VBN (VBN w/o relations) where explicit relations were not modeled directly.

\begin{table}[t]
\caption{Antonym learning evaluation.}\vspace{-4mm}
\label{hitrate-relations-table}
\begin{center}
\large
\begin{tabular}{lccc}
\toprule
\textsc{Model} & \textsc{HR@0.2\%} & \textsc{HR@1\%} & \textsc{MPR} \\
\midrule
VBN    & \textbf{3.8} & \textbf{12.4} & \textbf{92.5}\\
VBN w/o relations    & 2.1 & 8.3 & 76.1 \\
\bottomrule
\end{tabular}
\end{center}
\end{table}

\def \hfillx {\hspace*{-\textwidth} \hfill}

\begin{table}[t]
            \caption{Most similar words to \emph{lemma} ('Words' section) and most similar movies to \emph{Zoolander 2} ('Movies' section). \emph{TSLWM} stands for ``\emph{The Secret Life of Walter Mitty}''.}\vspace{-3mm}
        \label{lemma-table}
        \centering
        \large
        \begin{tabular}{ccc}
        \toprule
        \textsc{VBN} & \textsc{BSG} & \textsc{SG} \\
        \midrule
        \multicolumn{3}{c}{\text{\quad\quad\quad\qquad \textbf{Words} \quad\quad\quad}}\\
        \midrule
        \textcolor{mygreen}{\emph{theorem}}&   \emph{down} & \textcolor{red}{\emph{now}} \\ 
        \textcolor{mygreen}{\emph{subspace}}    & \emph{air} & \emph{already} \\
        \textcolor{mygreen}{\emph{polynomial}}   &  \emph{window} & \emph{still} \\ 
        \textcolor{mygreen}{\emph{operator}}  & \emph{water} & \emph{shot} \\
        \midrule
        \midrule
        \multicolumn{3}{c}{\text{\quad\quad\quad\qquad \textbf{Movies} \quad\quad\quad}}\\
        \midrule
        \textcolor{mygreen}{\emph{Zoolander 1}}     & \emph{Bossa Nova} & \emph{Manhunter} \\
        \textcolor{mygreen}{\emph{Tropic Thunder}}   & \emph{Escape from NY} & \textcolor{red}{\emph{TCM}}\\
        \textcolor{mygreen}{\emph{TSLWM}}   & \emph{Analyze That} & \emph{Scream 3} \\
        \textcolor{mygreen}{\emph{Meet the Fockers}}  & \emph{The Presidio} & \emph{8MM} \\
        
        \bottomrule
        \end{tabular}
\end{table}

\subsection{Qualitative Analysis}
In what follows, we provide a qualitative assessment of the contribution from each component of VBN.

{\bf \underline{Hierarchical Relations}:}
The word \emph{lemma} appears only \textbf{once} in our corpus, co-occurring only with the word \emph{now}. Thus, it exemplifies a rare word from the ``long-tail''. 
Table~\ref{lemma-table} (left) presents the four most similar words to \emph{lemma} suggested by SG (using the cosine similarity), BSG, and VBN (using Eq. \ref{eq:approximated-posterior-predicitive}).
According to WordNet, the parent of \emph{lemma} is \emph{proposition}, which is also the parent for \emph{theorem} (a more frequent word than \emph{lemma}). As can be seen in Tab.~\ref{lemma-table} ('Words' section), VBN suggests mathematical terms that better fit with \emph{lemma}, even though they never co-occurred with \emph{lemma} during training. 
Both BSG and SG, which do not utilize external information, suggest unrelated words. Moreover, the SG model is based on a point estimate solution, hence overconfident in its prediction, and ends up with (incorrectly) suggesting \emph{now} as the closest word to \emph{lemma}.

The same phenomenon is observed in the Movielens dataset. Here, only a single user watched both \emph{Zoolander 2} and \emph{The Texas Chainsaw Massacre (TCM)}. \emph{Zoolander 2} is a cold movie since it appears only once in the dataset (co-occurring only with \emph{TCM}). 
Further, note that \emph{Zoolander 2} was directed by Ben Stiller. In Tab.~\ref{lemma-table} (right), we see that all the recommended movies according to VBN are also Ben Stiller's movies, which exemplifies VBN's utilization of side information. 
Notably, the SG model promotes \emph{TCM} which is (arguably) not related to \emph{Zoolander 2}.

{\bf \underline{Bayesian Treatment}:}
Next, we show the model's ability to account for uncertainty in the target entity to be ranked.
Table~\ref{percent-table} depicts
the percentile rank (PR) of \emph{lemma} and \emph{Zoolander 2} w.r.t. \emph{now} and \emph{TCM}, respectively.  Note that in contrast to Table~\ref{lemma-table}, this time, the ``rare'' items are the target items.
Evidently, the PR in VBN and BSG is low (close to random ranking) due to the high variance of the target entities. SG is overconfident and ranks both irrelevant entities in the top $1\%$ and $4\%$, respectively.

{\bf \underline{Explicit Relations}:}
As detailed in Sec.~\ref{subsec:explicit}, the VBN explicitly models relations between entities. Following the experiment from Sec.~\ref{subsec:rel}, we present a qualitative example that demonstrates the merits of this capability. To this end, we consider the following pair of opposites (antonyms)- \emph{lowest} and \emph{highest}. We rank the words in the entire catalog w.r.t. \emph{lowest}, once according to $p(g^{opp}_{ij}=1|\bm{\mathcal{D}},\mathcal{H})$ (VBN) once according to $p(d_{ij}=1|\bm{\mathcal{D}},\mathcal{H})$ (which is computed based on the ablated version: VBN w/o relations).

Table~\ref{tab:rel-example} presents the top-$5$ words retrieved by both methods for the query \emph{lowest}. VBN ranks the word \emph{highest} in second place, while in the case of VBN w/o relations, \emph{highest} does not appear in the top~5 suggestions. Moreover, based on the co-occurrence data $d_{ij}$, VBN w/o relations undesirably promotes the word \emph{low}.

\begin{table}[t]
\caption{Top 5 words retrieved for the query \emph{lowest}.}\vspace{-4mm}
\label{tab:rel-example}
\begin{center}
\large
\begin{tabular}{cc}
\toprule
\textsc{VBN} & \textsc{VBN w/o relations} \\
\midrule
\emph{farther} & \emph{farther} \\
\textcolor{mygreen}{\emph{highest}} & \emph{method} \\
\emph{misgauged} & \emph{furthering} \\
\emph{landowners} & \textcolor{red}{\emph{low}} \\
\emph{witnessing} & \emph{denials}\\
\bottomrule
\end{tabular}
\end{center}
\end{table}

\begin{table}[t]
\caption{Percentile Rank (PR) of entity Y w.r.t. entity X.}\vspace{-4mm}
\label{percent-table}
\begin{center}
\large
\begin{tabular}{lccc}
\toprule
(X,Y) &\textsc{VBN} &  \textsc{BSG} & \textsc{SG} \\
\midrule
(\emph{now},\emph{lemma})&\textcolor{mygreen}{31.6}    &\textcolor{mygreen}{37.4} & \textcolor{red}{98.9} \\
(\emph{TCM}, \emph{Zoolander 2})&\textcolor{mygreen}{44.2}   & \textcolor{mygreen}{41.4} & \textcolor{red}{96.8} \\
\bottomrule
\end{tabular}
\end{center}
\end{table}

\section{Conclusion}
\label{sec:conclusion}
We presented VBN - a novel Bayesian model for effective representation learning in the long-tail, and small data scenarios. VBN introduces three complementary techniques: 1) Informative priors based on external hierarchical relations (e.g., taxonomy). 2) Explicit relational representations that enforce structure and consistency between entities that share a semantic relationship. 3) A tractable yet scalable VB optimization algorithm, followed by an analytical approximation to the posterior predictive integral, leading to a fast Bayesian inference. Extensive empirical evaluations on a variety of datasets and tasks show that VBN produces better representations than other methods in small data scenarios, and especially for long-tail entities.

In the future, we plan to expand the modeling of entities to multimodal distributions (e.g., Gaussian Mixture Models), and investigate the utilization of deep neural networks for incorporating further information sources (e.g., visual and audio signals) as priors.

\section{Broader Impact}
\label{sec:broder_impact}
Neural embedding methods have been introduced several years ago as a standard building block for many tasks in NLP and recommender systems, e.g., semantic similarity and collaborative filtering, etc. As such, these models play a substantial role in the AI revolution we are witnessing today.

Representation learning models are trained in a self-supervised fashion using large tabular (or textual) datasets. However, rare entities that suffer from insufficient statistics, are often poorly represented. As a result, the respective tasks often show inferior results in the ``long-tail''. One common mitigation, especially in language models, is to constantly increase training dataset size in the hope to gain more statistics for rare entities. However, aside from language models, this mitigation is mostly not applicable in other domains where datasets are limited and cannot be extended, e.g., in recommendation systems or medical informatics. Additionally, this challenge frequently arises when learning representations in propriety datasets.  

In this work, we proposed a Bayesian framework that allows utilizing side information (via external resources) and achieves superior representations for rare entities.
Encouraged by the model's results and our experience working with it, we genuinely predicate that this work pose a significant contribution within the scope of the aforementioned challenges. Therefore, we believe the NLP, recommender systems, medical informatics communities, and others can all benefit from the presented model,
and consider it as an elegant and effective way for modeling entities, 
especially when learning representations in the long-tail. Nevertheless, we humbly prefer to refrain from proclaiming broad declarations about its future ``societal consequences''. 

As to the ethical aspects of our work, we consider it similar to many other machine learning algorithms which can be seen as benign tools to be used for ``good'' or ``bad'' depending on the practitioner. 
As an exemplary use-case, we demonstrated successful entity representation learning of rare diagnoses that has the potential to be applied by health professionals to improve the well-being of others.
It is the opinion of the authors, that as a general rule, \emph{most} scientific contributions, such as the one made in this paper, have a positive impact in advancing humanity's technological achievements, even when the immediate impact is yet unknown.

\bibliography{references}

\begin{thebibliography}{10}

\bibitem{arora-etal-2020-learning}
Kushal Arora, Aishik Chakraborty, and Jackie C.~K. Cheung.
\newblock Learning lexical subspaces in a distributional vector space.
\newblock {\em Transactions of the Association for Computational Linguistics
  (TACL)}, pages 311--329, 2020.

\bibitem{barkan2017bayesian}
Oren Barkan.
\newblock Bayesian neural word embedding.
\newblock In {\em Thirty-First AAAI Conference on Artificial Intelligence},
  2017.

\bibitem{barkan2020attentive}
Oren Barkan, Avi Caciularu, Ori Katz, and Noam Koenigstein.
\newblock Attentive item2vec: Neural attentive user representations.
\newblock In {\em IEEE International Conference on Acoustics, Speech and Signal
  Processing (ICASSP)}, 2020.

\bibitem{barkan2020cold}
Oren Barkan, Avi Caciularu, Idan Rejwan, Ori Katz, Jonathan Weill, Itzik
  Malkiel, and Noam Koenigstein.
\newblock Cold item recommendations via hierarchical item2vec.
\newblock In {\em 2020 IEEE International Conference on Data Mining (ICDM)},
  pages 912--917. IEEE Computer Society, 2020.

\bibitem{barkan2020explainable}
Oren Barkan, Yonatan Fuchs, Avi Caciularu, and Noam Koenigstein.
\newblock Explainable recommendations via attentive multi-persona collaborative
  filtering.
\newblock In {\em ACM Conference on Recommender Systems (RecSys)}, 2020.

\bibitem{anchors}
Oren Barkan, Roy Hirsch, Ori Katz, Avi Caciularu, and Noam Koenigstein.
\newblock Anchor-based collaborative filtering.
\newblock In {\em Proceedings of the ACM International Conference on
  Information \& Knowledge Management (CIKM)}, 2021.

\bibitem{nam}
Oren Barkan, Ori Katz, and Noam Koenigstein.
\newblock Neural attentive multiview machines.
\newblock In {\em IEEE International Conference on Acoustics, Speech and Signal
  Processing (ICASSP)}, 2020.

\bibitem{i2v}
Oren {Barkan} and Noam {Koenigstein}.
\newblock Item2vec: Neural item embedding for collaborative filtering.
\newblock In {\em 2016 IEEE Machine Learning for Signal Processing (MLSP)},
  2016.

\bibitem{barkan2019cb2cf}
Oren Barkan, Noam Koenigstein, Eylon Yogev, and Ori Katz.
\newblock Cb2cf: a neural multiview content-to-collaborative filtering model
  for completely cold item recommendations.
\newblock In {\em Proceedings of the 13th ACM Conference on Recommender
  Systems}, pages 228--236, 2019.

\bibitem{barkan2020scalable}
Oren Barkan, Noam Razin, Itzik Malkiel, Ori Katz, Avi Caciularu, and Noam
  Koenigstein.
\newblock Scalable attentive sentence pair modeling via distilled sentence
  embedding.
\newblock In {\em Proceedings of the Conference on Artificial Intelligence
  (AAAI)}, 2020.

\bibitem{barkan-etal-2020-bayesian}
Oren Barkan, Idan Rejwan, Avi Caciularu, and Noam Koenigstein.
\newblock {B}ayesian hierarchical words representation learning.
\newblock In {\em Proceedings of the Annual Meeting of the Association for
  Computational Linguistics (ACL)}, 2020.

\bibitem{berry1941accuracy}
Andrew~C Berry.
\newblock The accuracy of the gaussian approximation to the sum of independent
  variates.
\newblock {\em Transactions of the American Mathematical Society},
  49(1):122--136, 1941.

\bibitem{bishop2006pattern}
Christopher~M Bishop.
\newblock {\em Pattern recognition and machine learning}.
\newblock springer, 2006.

\bibitem{bojanowski2017enriching}
Piotr Bojanowski, Edouard Grave, Armand Joulin, and Tomas Mikolov.
\newblock Enriching word vectors with subword information.
\newblock {\em Transactions of the Association for Computational Linguistics},
  5:135--146, 2017.

\bibitem{bollegala2016joint}
Danushka Bollegala, Mohammed Alsuhaibani, Takanori Maehara, and Ken-ichi
  Kawarabayashi.
\newblock Joint word representation learning using a corpus and a semantic
  lexicon.
\newblock In {\em Thirtieth AAAI Conference on Artificial Intelligence}, 2016.

\bibitem{bravzinskas2017embedding}
Arthur Bra{\v{z}}inskas, Serhii Havrylov, and Ivan Titov.
\newblock Embedding words as distributions with a bayesian skip-gram model.
\newblock {\em arXiv preprint arXiv:1711.11027}, 2017.

\bibitem{bruni2014multimodal}
Elia Bruni, Nam-Khanh Tran, and Marco Baroni.
\newblock Multimodal distributional semantics.
\newblock {\em Journal of Artificial Intelligence Research}, 49:1--47, 2014.

\bibitem{caciularu2021cross}
Avi Caciularu, Arman Cohan, Iz~Beltagy, Matthew~E Peters, Arie Cattan, and Ido
  Dagan.
\newblock Cross-document language modeling.
\newblock {\em arXiv preprint arXiv:2101.00406}, 2021.

\bibitem{caciularu-etal-2021-denoising}
Avi Caciularu, Ido Dagan, and Jacob Goldberger.
\newblock Denoising word embeddings by averaging in a shared space.
\newblock In {\em Proceedings of *SEM 2021: The Tenth Joint Conference on
  Lexical and Computational Semantics}, 2021.

\bibitem{devlin-etal-2019-bert}
Jacob Devlin, Ming-Wei Chang, Kenton Lee, and Kristina Toutanova.
\newblock {BERT}: Pre-training of deep bidirectional transformers for language
  understanding.
\newblock In {\em Proceedings of the 2019 Conference of the North {A}merican
  Chapter of the Association for Computational Linguistics: Human Language
  Technologies ({ACL})}, 2019.

\bibitem{faruqui-etal-2015-retrofitting}
Manaal Faruqui, Jesse Dodge, Sujay~Kumar Jauhar, Chris Dyer, Eduard Hovy, and
  Noah~A. Smith.
\newblock Retrofitting word vectors to semantic lexicons.
\newblock In {\em Proceedings of the 2015 Conference of the North {A}merican
  Chapter of the Association for Computational Linguistics: Human Language
  Technologies}, pages 1606--1615, 2015.

\bibitem{finkelstein2001placing}
Lev Finkelstein, Evgeniy Gabrilovich, Yossi Matias, Ehud Rivlin, Zach Solan,
  Gadi Wolfman, and Eytan Ruppin.
\newblock Placing search in context: The concept revisited.
\newblock In {\em Proceedings of the 10th international conference on World
  Wide Web}, pages 406--414, 2001.

\bibitem{harper2015movielens}
F~Maxwell Harper and Joseph~A Konstan.
\newblock The movielens datasets: History and context.
\newblock {\em Acm transactions on interactive intelligent systems (tiis)},
  5(4):1--19, 2015.

\bibitem{3061142}
Ruining He, Chunbin Lin, Jianguo Wang, and Julian McAuley.
\newblock Sherlock: Sparse hierarchical embeddings for visually-aware one-class
  collaborative filtering.
\newblock In {\em Proceedings of the International Joint Conference on
  Artificial Intelligence ({IJCAI})}, 2016.

\bibitem{hill2015simlex}
Felix Hill, Roi Reichart, and Anna Korhonen.
\newblock Simlex-999: Evaluating semantic models with (genuine) similarity
  estimation.
\newblock {\em Computational Linguistics}, 41(4):665--695, 2015.

\bibitem{hice2019}
Ziniu Hu, Ting Chen, Kai-Wei Chang, and Yizhou Sun.
\newblock Few-shot representation learning for out-of-vocabulary words.
\newblock In {\em Proceedings of the 57th Annual Meeting of the Association for
  Computational Linguistics, ({ACL})}, 2019.

\bibitem{huang2012improving}
Eric~H Huang, Richard Socher, Christopher~D Manning, and Andrew~Y Ng.
\newblock Improving word representations via global context and multiple word
  prototypes.
\newblock In {\em Proceedings of the 50th Annual Meeting of the Association for
  Computational Linguistics: Long Papers-Volume 1}, pages 873--882. Association
  for Computational Linguistics, 2012.

\bibitem{huang-etal-2019-glossbert}
Luyao Huang, Chi Sun, Xipeng Qiu, and Xuanjing Huang.
\newblock {G}loss{BERT}: {BERT} for word sense disambiguation with gloss
  knowledge.
\newblock In {\em Proceedings of the 2019 Conference on Empirical Methods in
  Natural Language Processing and the 9th International Joint Conference on
  Natural Language Processing (EMNLP-IJCNLP)}, pages 3507--3512, 2019.

\bibitem{jaakkola1997variational}
Tommi Jaakkola and Michael Jordan.
\newblock A variational approach to bayesian logistic regression models and
  their extensions.
\newblock In {\em Sixth International Workshop on Artificial Intelligence and
  Statistics}, volume~82, 1997.

\bibitem{johnson2016mimic}
Alistair~EW Johnson, Tom~J Pollard, Lu~Shen, H~Lehman Li-wei, Mengling Feng,
  Mohammad Ghassemi, Benjamin Moody, Peter Szolovits, Leo~Anthony Celi, and
  Roger~G Mark.
\newblock Mimic-iii, a freely accessible critical care database.
\newblock {\em Scientific data}, 3:160035, 2016.

\bibitem{yahoo2008music}
Noam Koenigstein, Gideon Dror, and Yehuda Koren.
\newblock Yahoo! music recommendations: modeling music ratings with temporal
  dynamics and item taxonomy.
\newblock In {\em Proceedings of the fifth ACM conference on Recommender
  systems}, pages 165--172, 2011.

\bibitem{koenigstein13}
Noam Koenigstein and Ulrich Paquet.
\newblock Xbox movies recommendations: Variational bayes matrix factorization
  with embedded feature selection.
\newblock In {\em Proceedings of the 7th ACM Conference on Recommender
  Systems}, RecSys '13, page 129–136. Association for Computing Machinery,
  2013.

\bibitem{Koren_MF}
Y.~{Koren}, R.~{Bell}, and C.~{Volinsky}.
\newblock Matrix factorization techniques for recommender systems.
\newblock {\em Computer}, 42(8):30--37, 2009.

\bibitem{levine2019sensebert}
Yoav Levine, Barak Lenz, Or~Dagan, Dan Padnos, Or~Sharir, Shai Shalev-Shwartz,
  Amnon Shashua, and Yoav Shoham.
\newblock {SenseBERT}: Driving some sense into {BERT}.
\newblock {\em arXiv preprint arXiv:1908.05646}, 2019.

\bibitem{cvae}
Xiaopeng Li and James She.
\newblock Collaborative variational autoencoder for recommender systems.
\newblock In {\em Proceedings of the ACM SIGKDD International Conference on
  Knowledge Discovery and Data Mining (KDD)}, 2017.

\bibitem{li-etal-2016-joint}
Yuezhang Li, Ronghuo Zheng, Tian Tian, Zhiting Hu, Rahul Iyer, and Katia
  Sycara.
\newblock Joint embedding of hierarchical categories and entities for concept
  categorization and dataless classification.
\newblock In {\em Proceedings of {COLING} 2016, the 26th International
  Conference on Computational Linguistics: Technical Papers}, pages 2678--2688,
  2016.

\bibitem{3186150}
Dawen Liang, Rahul~G. Krishnan, Matthew~D. Hoffman, and Tony Jebara.
\newblock Variational autoencoders for collaborative filtering.
\newblock In {\em Proceedings of the World Wide Web Conference ({WWW})}, 2018.

\bibitem{lin2019projection}
Chin Lin, Yu-Sheng Lou, Dung-Jang Tsai, Chia-Cheng Lee, Chia-Jung Hsu,
  Ding-Chung Wu, Mei-Chuen Wang, and Wen-Hui Fang.
\newblock Projection word embedding model with hybrid sampling training for
  classifying icd-10-cm codes: longitudinal observational study.
\newblock {\em JMIR medical informatics}, 7(3):e14499, 2019.

\bibitem{liu2019roberta}
Yinhan Liu, Myle Ott, Naman Goyal, Jingfei Du, Mandar Joshi, Danqi Chen, Omer
  Levy, Mike Lewis, Luke Zettlemoyer, and Veselin Stoyanov.
\newblock Roberta: A robustly optimized bert pretraining approach.
\newblock {\em arXiv preprint arXiv:1907.11692}, 2019.

\bibitem{luong2013better}
Thang Luong, Richard Socher, and Christopher Manning.
\newblock Better word representations with recursive neural networks for
  morphology.
\newblock In {\em Proceedings of the Seventeenth Conference on Computational
  Natural Language Learning}, pages 104--113, 2013.

\bibitem{mackay1992evidence}
David~JC MacKay.
\newblock The evidence framework applied to classification networks.
\newblock {\em Neural computation}, 4(5):720--736, 1992.

\bibitem{mikolov2013distributed}
Tomas Mikolov, Ilya Sutskever, Kai Chen, Greg~S Corrado, and Jeff Dean.
\newblock Distributed representations of words and phrases and their
  compositionality.
\newblock In {\em Advances in neural information processing systems}, pages
  3111--3119, 2013.

\bibitem{miller1995wordnet}
George~A Miller.
\newblock {WordNet}: a lexical database for english.
\newblock {\em Communications of the ACM}, 38(11):39--41, 1995.

\bibitem{miller1990introduction}
George~A Miller, Richard Beckwith, Christiane Fellbaum, Derek Gross, and
  Katherine~J Miller.
\newblock Introduction to {WordNet}: An on-line lexical database.
\newblock {\em International journal of lexicography}, 3(4):235--244, 1990.

\bibitem{miller1993semantic}
George~A Miller, Claudia Leacock, Randee Tengi, and Ross~T Bunker.
\newblock A semantic concordance.
\newblock In {\em Proceedings of the workshop on Human Language Technology},
  pages 303--308. Association for Computational Linguistics, 1993.

\bibitem{3327687}
Boris Muzellec and Marco Cuturi.
\newblock Generalizing point embeddings using the wasserstein space of
  elliptical distributions.
\newblock In {\em Proceedings of the International Conference on Neural
  Information Processing Systems ({NIPS})}, 2018.

\bibitem{paquet2013}
Ulrich Paquet and Noam Koenigstein.
\newblock One-class collaborative filtering with random graphs.
\newblock In {\em Proceedings of the 22nd International Conference on World
  Wide Web}, page 999–1008, 2013.

\bibitem{pennington2014glove}
Jeffrey Pennington, Richard Socher, and Christopher Manning.
\newblock Glove: Global vectors for word representation.
\newblock In {\em Proceedings of the 2014 conference on empirical methods in
  natural language processing (EMNLP)}, pages 1532--1543, 2014.

\bibitem{Peters2019KnowledgeEC}
Matthew~E. Peters, Mark Neumann, Robert~L Logan, Roy Schwartz, Vidur Joshi,
  Sameer Singh, and Noah~A. Smith.
\newblock Knowledge enhanced contextual word representations.
\newblock In {\em EMNLP}, 2019.

\bibitem{pinter-etal-2017-mimicking}
Yuval Pinter, Robert Guthrie, and Jacob Eisenstein.
\newblock Mimicking word embeddings using subword {RNN}s.
\newblock In {\em Proceedings of the Conference on Empirical Methods in Natural
  Language Processing ({EMNLP})}, 2017.

\bibitem{salakhutdinov2008bayesian}
Ruslan Salakhutdinov and Andriy Mnih.
\newblock Bayesian probabilistic matrix factorization using markov chain monte
  carlo.
\newblock In {\em Proceedings of the 25th international conference on Machine
  learning}, pages 880--887. ACM, 2008.

\bibitem{schick-schutze-2019-attentive}
Timo Schick and Hinrich Sch{\"u}tze.
\newblock Attentive mimicking: Better word embeddings by attending to
  informative contexts.
\newblock In {\em Proceedings of the 2019 Conference of the North {A}merican
  Chapter of the Association for Computational Linguistics: Human Language
  Technologies ({NAACL})}, 2019.

\bibitem{schick2020rare}
Timo Schick and Hinrich Sch{\"u}tze.
\newblock Rare words: A major problem for contextualized representation and how
  to fix it by attentive mimicking.
\newblock In {\em Proceedings of the Thirty-Fourth AAAI Conference on
  Artificial Intelligence}, 2020.

\bibitem{tifrea2018poincare}
Alexandru Tifrea, Gary Becigneul, and Octavian-Eugen Ganea.
\newblock Poincare glove: Hyperbolic word embeddings.
\newblock In {\em International Conference on Learning Representations}, 2019.

\bibitem{vilnis2014word}
Luke Vilnis and Andrew McCallum.
\newblock Word representations via gaussian embedding.
\newblock {\em International Conference on Learning Representations}, 2015.

\bibitem{wang2019neural}
Xiang Wang, Xiangnan He, Meng Wang, Fuli Feng, and Tat-Seng Chua.
\newblock Neural graph collaborative filtering.
\newblock In {\em Proceedings of the international ACM conference on Research
  and development in Information Retrieval (SIGIR)}, 2019.

\bibitem{wu2016collaborative}
Yao Wu, Christopher DuBois, Alice~X Zheng, and Martin Ester.
\newblock Collaborative denoising auto-encoders for top-n recommender systems.
\newblock In {\em Proceedings of the ACM International Conference on Web Search
  and Data Mining (WSDM)}, 2016.

\bibitem{yu-dredze-2014-improving}
Mo~Yu and Mark Dredze.
\newblock Improving lexical embeddings with semantic knowledge.
\newblock In {\em Proceedings of the Annual Meeting of the Association for
  Computational Linguistics (ACL)}. Association for Computational Linguistics,
  2014.

\bibitem{zhang-etal-2014-word}
Jingwei Zhang, Jeremy Salwen, Michael Glass, and Alfio Gliozzo.
\newblock Word semantic representations using {B}ayesian probabilistic tensor
  factorization.
\newblock In {\em Proceedings of the 2014 Conference on Empirical Methods in
  Natural Language Processing ({EMNLP})}, pages 1522--1531, 2014.

\bibitem{10.1145/3383313.3412239}
Yin Zhang, Ziwei Zhu, Yun He, and James Caverlee.
\newblock Content-collaborative disentanglement representation learning for
  enhanced recommendation.
\newblock In {\em ACM Conference on Recommender Systems (RecSys)}, 2020.

\end{thebibliography}
\bibliographystyle{plain}

\end{document}